\documentclass[letterpaper,conference]{IEEEtran}
\IEEEoverridecommandlockouts
\usepackage{cite}
\usepackage{amsmath,amssymb,amsfonts}
\usepackage{algorithmic}
\usepackage{graphicx}
\usepackage{textcomp}

\usepackage{xcolor}
\usepackage{multirow}
\usepackage{booktabs}
\usepackage{url}
\usepackage{subfigure}
\usepackage[colorlinks=true,linkcolor=blue,citecolor=blue]{hyperref}

\def\BibTeX{{\rm B\kern-.05em{\sc i\kern-.025em b}\kern-.08em
    T\kern-.1667em\lower.7ex\hbox{E}\kern-.125emX}}
\begin{document}

\title{\LARGE \textbf{ViT-DD: Multi-Task Vision Transformer for Semi-Supervised Driver Distraction Detection
}}

\author{
Yunsheng Ma
and Ziran Wang
\thanks{Y. Ma and Z. Wang are with College of Engineering, Purdue University, West Lafayette, IN 47907, USA. Emails: {\tt\small \{yunsheng, ziran\}@purdue.edu}}
}

\maketitle

\begin{abstract}
Ensuring traffic safety and mitigating accidents in modern driving is of paramount importance, and computer vision technologies have the potential to significantly contribute to this goal. This paper presents a multi-modal Vision Transformer for Driver Distraction Detection (termed ViT-DD), which incorporates inductive information from training signals related to both distraction detection and driver emotion recognition. Additionally, a self-learning algorithm is developed, allowing for the seamless integration of driver data without emotion labels into the multi-task training process of ViT-DD. Experimental results reveal that the proposed ViT-DD surpasses existing state-of-the-art methods for driver distraction detection by 6.5\% and 0.9\% on the SFDDD and AUCDD datasets, respectively. 
\end{abstract}

\newcommand{\etal}{et al.}

\section{Introduction}
According to National Highway Traffic Safety Administration
(NHTSA), there were 38,824 people killed in motor vehicle crashes on U.S. roadways during 2020. Among these cases, 3,142 (or 8.1\%) are distraction-affected crashes, i.e., a crash involving at least one driver who was distracted. Distracted driving is defined by NHTSA as any activity that diverts attention away from safe driving, such as talking or texting on a cell phone, eating and drinking, chatting with others in the car, and fiddling with the audio, entertainment, or navigation system~\cite{NHTSA2022overview}.

During the past decade, rapid development has been witnessed worldwide in intelligent vehicle technology, where advancements in perception~\cite{lindenmaier_object-level_2023,wu_surround-view_2023,li_learning_2023,meng_hydro-3d_2023}, communication~\cite{chapala_intelligent_2023,gao_network-induced_2023,lin_cader_2023,zhang_memory-anticipation_2023,chen_robustly_2023,eskandari_slaps_2023,wang_conflict_2023}, and computation introduced numerous emerging applications on intelligent vehicles~\cite{zhang_ad4che_2023,farooq_evaluation_2023,wu_uncertainty-aware_2023,zhou_acp-based_2023,liu_software-defined_2023}. As a core element of intelligent vehicles, driving automation systems, such as Advanced Driver-Assistance Systems (ADAS) and Automated Driving Systems (ADS), have been designed to support human drivers either by providing warnings to reduce risk exposure, or by assisting the vehicle actuation to relieve drivers’ burden on some of the driving tasks~\cite{farooq_evaluation_2023,li_development_2023}. When functioning, these systems can help the driver safely navigate the vehicle through tricky traffic scenarios when he/she is distracted by some other tasks. 

However, a driver can also over-trust the driving automation system, especially when the system is categorized as SAE Level 3 (i.e., conditional driving automation): When the automated driving features are engaged, the driver is allowed to take his/her hands off the steering wheel and feet off the pedals, but he/she needs to stay alert and get ready to take over the driving task when the system requests \cite{SAELevel}. Due to human nature, the attention from the driver on road conditions can get diminished when he/she is not in charge of driving, and the involvement of distracted behaviors can decrease the driver's capability of taking over, which in turn leads to traffic accidents.

It can be envisioned in future transportation systems that intelligent vehicles can detect and identify driver distractions, then warn the driver against them or take precautionary measures. Therefore, in this paper, a multi-modal Vision Transformer (termed ViT-DD) is proposed to exploit inductive information contained in the training signals of both emotion recognition and distraction detection, along with a novel pseudo-labeled multi-task training algorithm that leverages the knowledge in an independent emotion recognition teacher model to train a student ViT-DD.

In summary, the contributions of this paper are threefold:

\begin{itemize}
    \item This paper explores the detection of driver distractions using a pure Transformer-based architecture and a semi-supervised learning setup, which, to the best of the authors' knowledge, has not been previously investigated.
    \item A multi-modal Vision Transformer (ViT-DD) is developed for driver distraction detection. The proposed ViT-DD model leverages a novel semi-supervised learning approach to incorporate driver data without emotion labels in the multi-task training process.
    \item Extensive experiments are conducted on two benchmark datasets (SFDDD and AUCDD), and the results show that the proposed methodology outperforms the state-of-the-art approaches.
\end{itemize}

\section{Background} \label{sec:Related}

\subsection{Vision Transformer}
Since AlexNet~\cite{krizhevsky2017imagenet}, convolutional neural networks (CNNs) have been the dominant methodology for learning visual representations of images in computer vision (CV)\cite{Simonyan2015VeryDC,he2016deep}. Vision Transformer (ViT)~\cite{visiontransformer}, on the other hand, has recently achieved state-of-the-art performances on a variety of CV tasks and garnered significant interest from the CV community~\cite{cui_survey_2024,ma_macp_2024}. 

ViT seldom employs convolution kernels (i.e. the core of CNNs). Instead, it relies on the self-attention mechanism~\cite{transformer} to provide context information for input visual tokens, which is inspired by tasks in natural language processing. In addition to the initial patch extraction process, ViT does not introduce image-specific inductive biases into its architecture. 

There have been some attempts to apply Transformer architecture to the detection of driver distractions. For example, DDR-ViT~\cite{9661254} is a distracted driving recognition model based on a fine-tuned Vision Transformer, proposed to solve the problem of accurate and timely recognition of driving behavior. TransDARC~\cite{peng2022transdarc} presented a vision-based framework for recognizing secondary driver behaviors based on visual transformers and an additional augmented feature distribution calibration module. However, none of these works consider driver emotion as an additional modality that can significantly improve performance in detecting distracted behaviors.


\subsection{Multi-Task Learning and Self-Training}
Multi-Task Learning (MTL) is an inductive transfer mechanism with the primary objective of enhancing generalization performance. 
Learning one task at a time is the standard for machine learning. However, Caruana\cite{caruana1997multitask} contends that this strategy is sometimes ineffective since it disregards a potentially rich source of information accessible in many real-world problems, i.e., the information contained in the training signals of other tasks drawn from the same domain. If the tasks can share what they learn, it may be preferable to require the learner to learn many capabilities simultaneously.
In computer vision, a popular method for MTL is to employ a single encoder to learn a shared representation, followed by numerous task-specific decoders \cite{cui_redformer_2023}. In this paper, a similar strategy is employed by training one main backbone model together with several small task-specific heads.

Self-training is an approach for incorporating unlabeled data into a supervised learning task~\cite{liu_self-supervised_2023}.
It is one of the earliest semi-supervised learning approaches, which generates pseudo labels for unlabeled data using a supervised model. Recently, Ghiasi \etal \cite{ghiasi2021multi} proposes multi-task self-training (MuST), an approach for generating generalized visual representations using multi-task learning with pseudo labels. This method differs from the approach presented in this paper in that the multi-task learning strategy is employed for both output and various input modalities.

\subsection{Facial Expression Recognition and Driver Distraction Detection}
Facial expression recognition (FER) is an image classification problem that recognizes the emotion state of individuals~\cite{zhao2020end}. AffectNet-7 \cite{mollahosseini2017affectnet} is currently the largest publicly available FER dataset, which comprises images labeled with Ekman's six fundamental emotions \cite{ekman1992argument}, namely \textit{happy, sad, surprise, fear, disgust,} and \textit{anger}, plus an additional \textit{neutral} category. 
In this paper, FER is employed to drivers' face images to evaluate his or her emotion state, therefore acquiring additional information to detect driver distractions through multi-task learning.

\section{Methodology} \label{sec:Method}
\begin{figure*}[t]
    \centering
    \includegraphics[width=\textwidth]{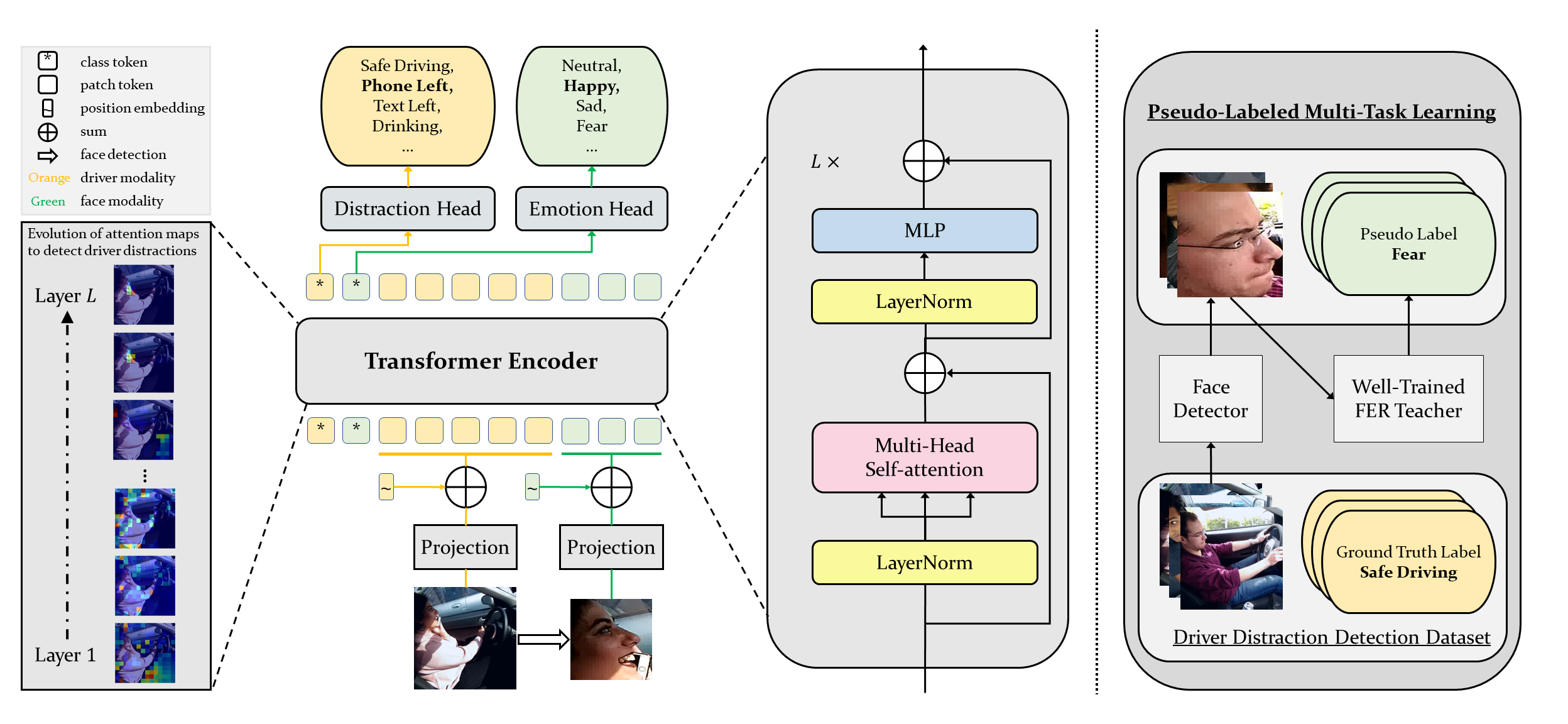}
    \caption{\textbf{(Left) The framework of the proposed ViT-DD:} First, a face detector is applied to the input signal from an in-cabin camera to acquire the driver's facial area. Then, the driver and face images are divided into patches and independently embedded into visual tokens. Next, the driver and face embeddings are added with their respective position embeddings, and the resulting sequence is concatenated. In addition, tokens representing distractions and emotions are prepended. The sequence of class and visual tokens are then iteratively updated through $L$ Transformer layers. The class tokens from the final sequence are used to recognize the driver's distraction and emotion states through their corresponding MLP heads. \textbf{(Right) Pseudo-labeled Multi-Task Learning:} A well-trained Facial Expression Recognition Teacher ViT is employed to label the unlabeled drivers' face images in order to create a multi-task driver dataset. The dataset containing both ground-truth distraction labels and pseudo emotion labels is then applied to train a student ViT-DD model with multi-task learning.}
    \label{fig:vit-dd}
\end{figure*}

In this section, a novel multi-task ViT for semi-supervised driver distraction detection is proposed, where the overall framework is shown in Fig. \ref{fig:vit-dd}. Specifically, ViT-DD has two input modalities, i.e. driver and face, to exploit information contained in the training signals of both distraction detection and emotion recognition. The input images from both modalities are separated into patches, linearly projected to fixed-dimensional visual tokens, and encoded using a Transformer encoder. Task-specific classification heads are applied to the output sequence of the Transformer encoder to generate the prediction results. The training of ViT-DD is conducted through a novel multi-task multi-modal self-training technique.

\subsection{Model Overview}
\newcommand{\x}{\mathbf{x}}
\newcommand{\y}{\mathbf{y}}
\newcommand{\ttt}{\mathbf{t}}
\newcommand{\ttti}{\ttt^{(i)}}
\newcommand{\xxi}{\x^{(i)}}
\newcommand{\z}{\mathbf{z}}
\newcommand{\q}{\mathbf{q}}
\newcommand{\kk}{\mathbf{k}}
\newcommand{\vv}{\mathbf{v}}
\newcommand{\vvi}{\vv^{(i)}}
\newcommand{\Ei}{E^{(i)}}
\newcommand{\X}{\mathcal{X}}
\newcommand{\Y}{\mathcal{Y}}

The backbone of ViT-DD is a ViT \cite{visiontransformer}, with different patch projection layers applied to each input modality (driver and face). 
Specially, the input space for each modality is defined by $\X^{(i)}$, where $i\in\left\{0,1\right\}$.
The input image $\x^{(i)}\in\X^{(i)}\subseteq\mathbb{R}^{C\times H_i\times W_i}$ from modality $i$ is sliced into patches and then flattened to $\vv^{(i)}\in\mathbb{R}^{N_i\times(P^2\cdot C)}$, where $(P,P)$ is the patch size, $C$ is the number of channels of the input image, and $N_i=H_iW_i/P^2$ is the number of patches for each modality. Next, the flattened patches are linear projected to $D$ dimensional tokens with the projection matrix $E^{(i)}\in\mathbb{R}^{(P^2\cdot C)\times D}$, followed by position embedding $E_{\text{pos}}\in\mathbb{R}^{N_i\times D}$. 
Additionally, class tokens ($\ttt^{(i)}_{\text{class}}$) for each modality with a learnable embedding are prepended to the beginning of the input sequence. All input tokens are then concatenated into a combined sequence $\z_0$ (Eq.\ref{eq:z0}) and sent to the same Transformer Encoder as input. 

\begin{align}
\Bar{\x}^{(i)}&=[\ttti_1 \Ei;\cdots;\ttti_{N_i} \Ei] + \Ei_{\text{pos}}, &i=0,1 \\
z^0 &= [\ttt^{(0)}_{\text{class}};\ttt^{(1)}_{\text{class}};\Bar{\x}^{(0)},\Bar{\x}^{(1)}] \label{eq:z0}\\
\z'_\ell &= \operatorname{MSA}(\operatorname{LN}(\z_{\ell-1})) + \z_{\ell-1}, &\ell=1\ldots L\\
\z_\ell &= \operatorname{MLP}(\operatorname{LN}(\z'_\ell)) + \z'_\ell, &\ell=1\ldots L\\
\y^{(i)} &= \operatorname{LN}(\z_L^i), &i=0,1 \label{eq:head}
\end{align}

The Transformer encoder then learns the fused driver behavior and emotion representation by stacking $L$ Transformer blocks . A Multilayer Perceptron (MLP) module and a Multihead Self-attention (MSA) module are included in each block. Additionally, LayerNorm (LN)~\cite{ba2016layernrom} is also adopted prior to each module. Self-attention (SA) is the key component of Transformer blocks, in which, the input vector $\z_{\ell-1}$ is first transformed into three separate vectors: the query vector $\q$, the key vector $\kk$, and the value vector $\vv$, all of the same dimension $\q,\kk,\vv\in\mathbb{R}^D$. After that, the attention scores are constructed by the following function:
\begin{equation}
    \operatorname{SA}(\z_{\ell-1}) = \operatorname{softmax}\left(\frac{W_Q\q\cdot(W_K \kk)^\top}{\sqrt{D_H}}\right)\left(W_V \vv\right)
\end{equation}
where $W_Q, W_K, W_V \in \mathbb{R}^{D_H\times D}$ are learnable parameters of three linear projections and $D_H=D$. 

MSA is an extension of SA with $H$ self-attention heads, which can be formulated as:
\begin{equation}
    \operatorname{MSA}(\z_{\ell-1})=W_P\left[\operatorname{SA}_1(\z_{\ell-1});\cdots;\operatorname{SA}_H(\z_{\ell-1})\right]
\end{equation}
where $W_P\in\mathbb{R}^{D\times (H\cdot D_H)}$, and $D_H$ is typically set to $D/H$. To obtain the final multi-task prediction probabilities of the driver's distractions and emotions, the states of the \textit{class} tokens at the output of the Transformer encoder ($\z_L^i$) are fed into the respective classification heads, with the standard cross-entropy loss adopted. The final loss is the weighed sum of the loss of each head.

For all experiments, a ViT-B \cite{visiontransformer} pretrained on ImageNet\cite{imagenet} with a patch size of $16\times 16$ pixels is employed as the backbone. Specially, the latent vector size $D$ is 768, the patch size $P$ is 16, the layer depth $L$ is 12, and the number of attention heads $H$ is 12. 

\subsection{Pseudo-Labeled Multi-Task Training}
Pseudo labeling offers the benefit of not requiring a large multi-task dataset with all required labels. Having access to a well-trained neural network, that can produce pseudo labels of other domains on the dataset we wish to work on, can be effective. Pseudo labeling is a one-time preprocessing method applicable to RGB datasets of variable size. Compared to the training cost, this phase is computationally inexpensive \cite{bachmann2022multimae}. 

The proposed multi-task multi-modal self-training algorithm has four steps. 
First a teacher ViT is trained on AffectNet-7 \cite{mollahosseini2017affectnet}, a large facial emotion recognition dataset, to enable it to recognize the facial expressions of drivers.  
Second, RetinaFace\cite{RetinaFace}, a face detector, is used to detect and crop face images in the driver distraction detection datasets. 
Next, the FER teacher model is used to label the unlabeled drivers' face images to create a multi-task pseudo-labeled driver dataset. 
Finally, the driver dataset, which now contains both supervised labels for distraction detection and pseudo labels from the teacher model for emotion recognition, is then employed to train a student ViT-DD model with multi-task multi-modal learning. To manage the situation in which the driver's face cannot be detected, an additional \textit{Non-Face} label is added to the emotion classification task, and in such case, a blank image is fed to the face input.

\newcommand{\vspacefigtwo}{\vspace{0.5mm}}
\newcommand{\fontsizefigtwo}{\scriptsize}
\newcommand{\sizeafigtwo}{0.14}
\newcommand{\sizebfigtwo}{0.07}
\newcommand{\sizelfigtwo}{0.19}
\newcommand{\sizerfigtwo}{0.80}

\begin{figure*}[!t]%
\begin{minipage}{\textwidth}
\begin{minipage}{\sizelfigtwo\textwidth}
\centering
\fontsizefigtwo
\includegraphics[width=0.8\linewidth]{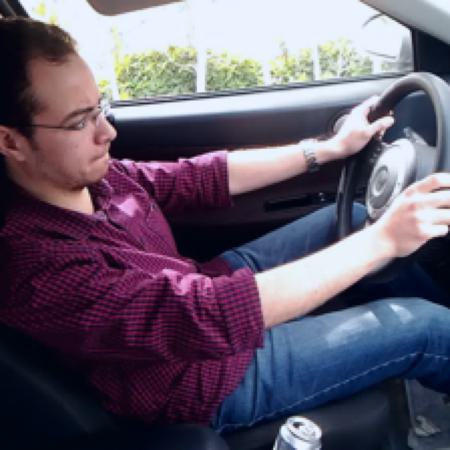}\\
Safe Driving
\end{minipage}
\begin{minipage}{\sizerfigtwo\textwidth}
\centering
\begin{minipage}{\textwidth}
\centering
\fontsizefigtwo
\includegraphics[width=\sizeafigtwo\linewidth,height=\sizeafigtwo\linewidth]{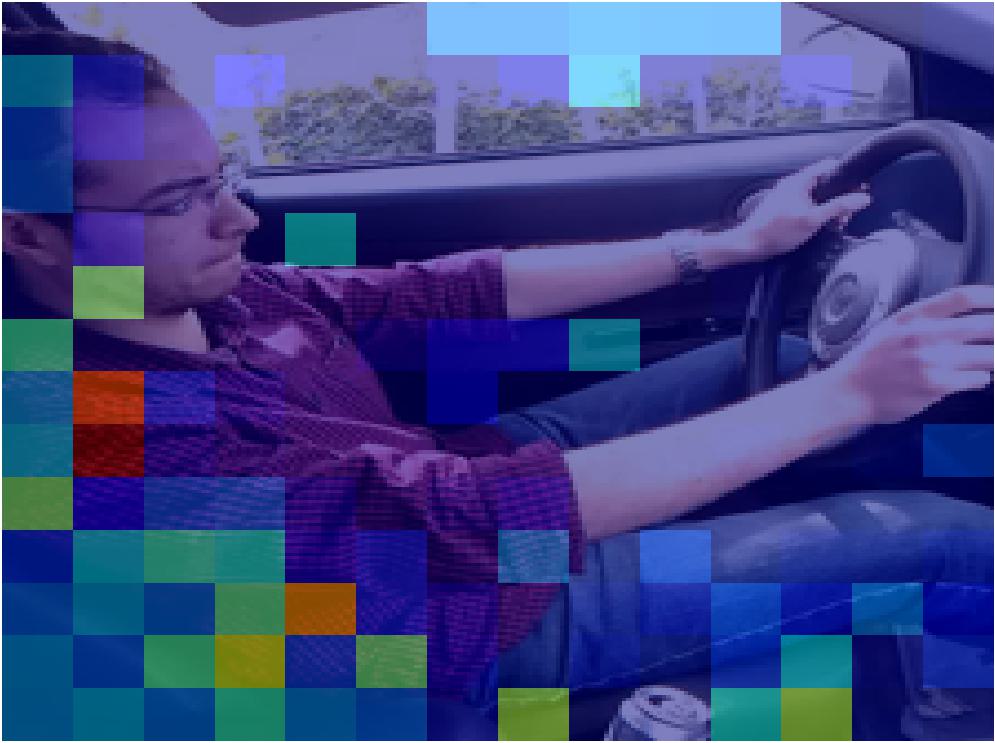}
\includegraphics[width=\sizeafigtwo\linewidth,height=\sizeafigtwo\linewidth]{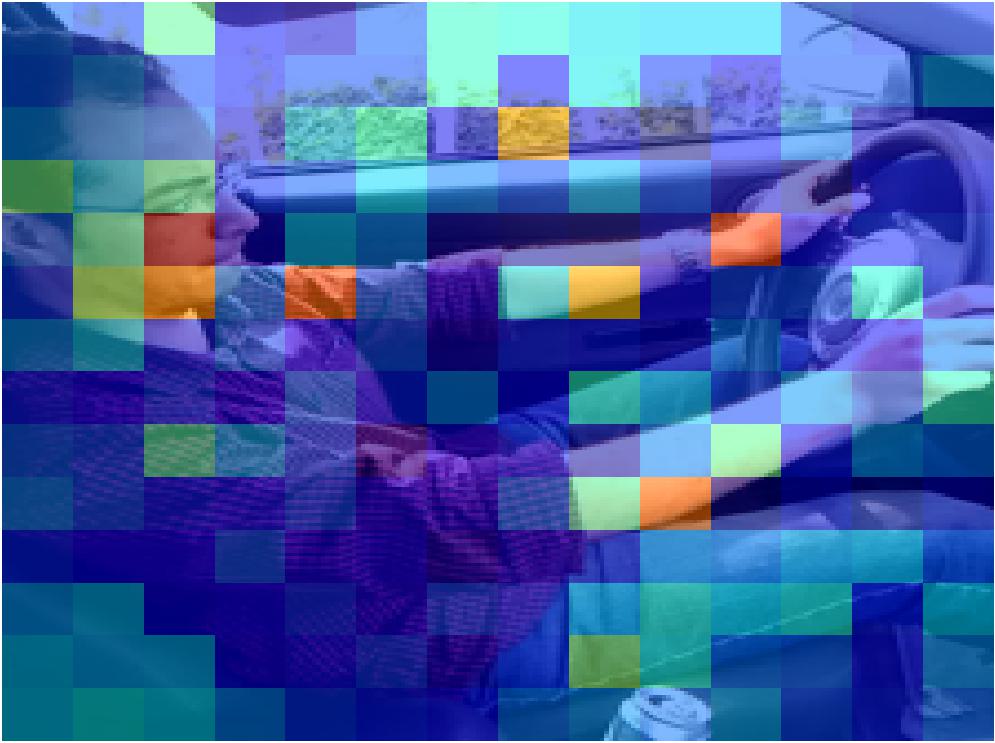}
\includegraphics[width=\sizeafigtwo\linewidth,height=\sizeafigtwo\linewidth]{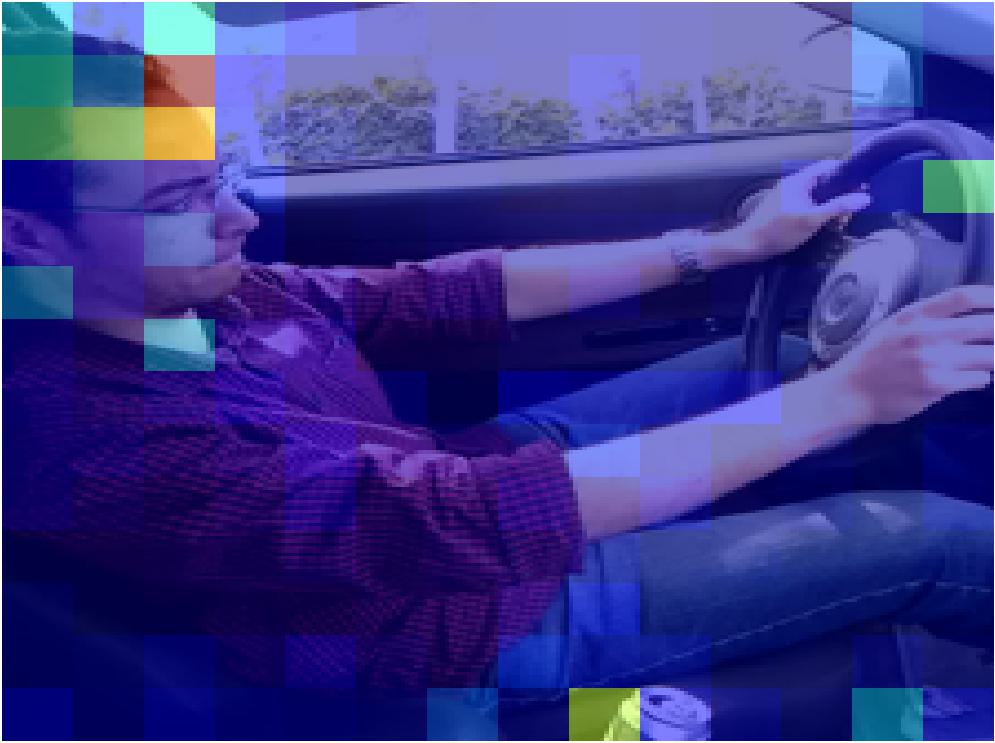}
\includegraphics[width=\sizeafigtwo\linewidth,height=\sizeafigtwo\linewidth]{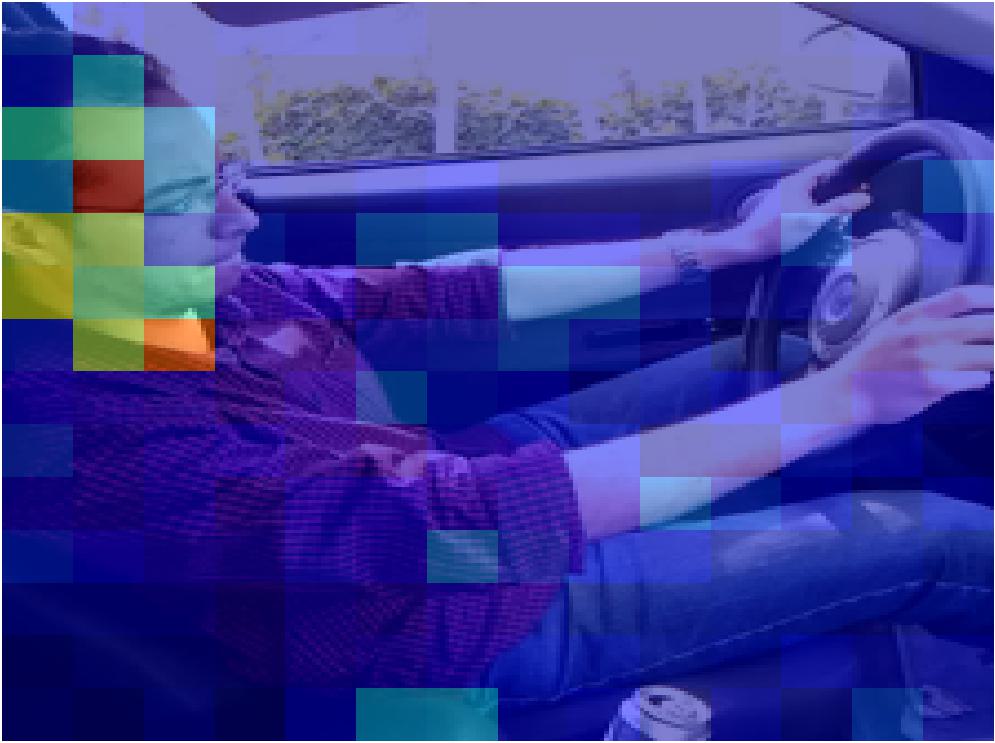}
\includegraphics[width=\sizeafigtwo\linewidth,height=\sizeafigtwo\linewidth]{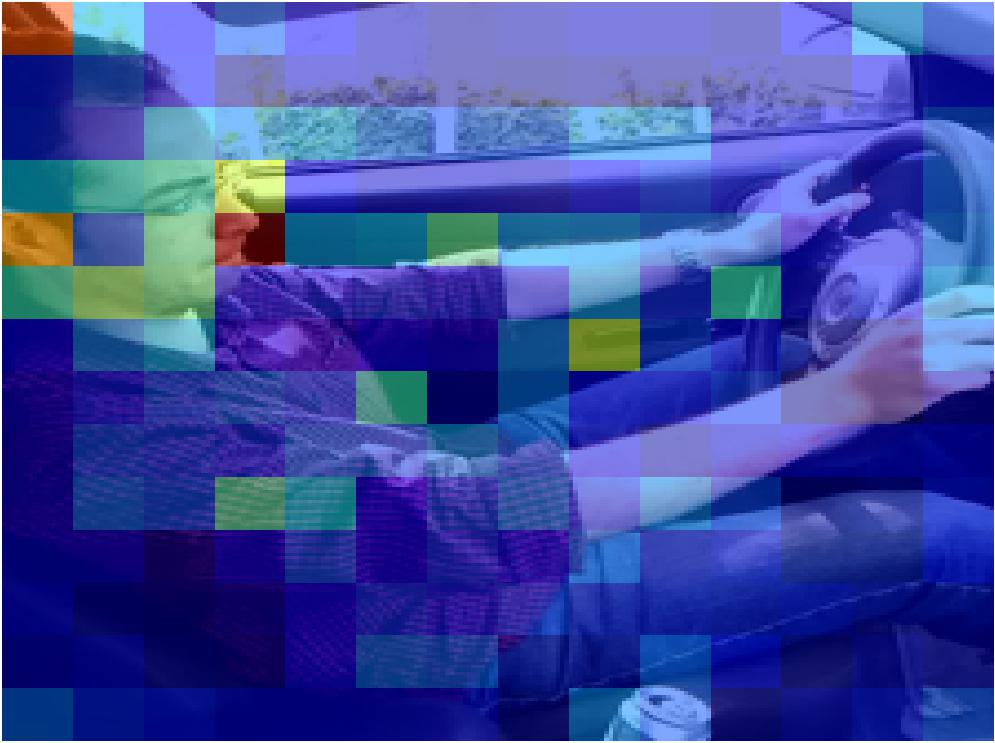} 
\includegraphics[width=\sizeafigtwo\linewidth,height=\sizeafigtwo\linewidth]{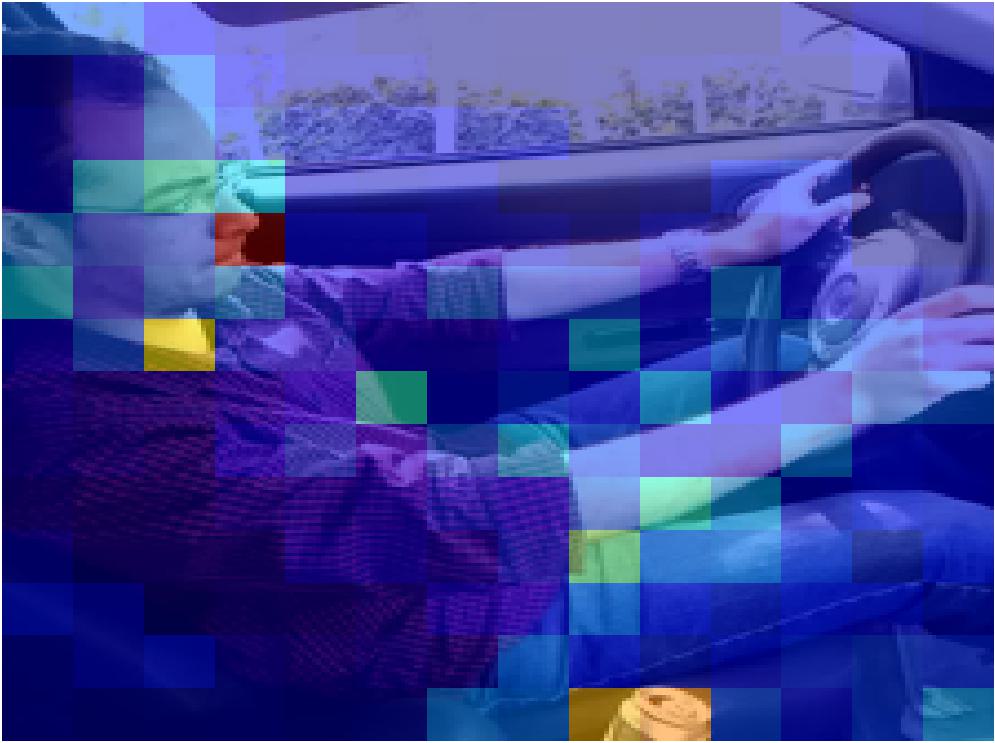} \\
Layer 1 to Layer 6
\vspacefigtwo
\end{minipage}
\begin{minipage}{\textwidth}
\centering
\fontsizefigtwo
\includegraphics[width=\sizeafigtwo\linewidth,height=\sizeafigtwo\linewidth]{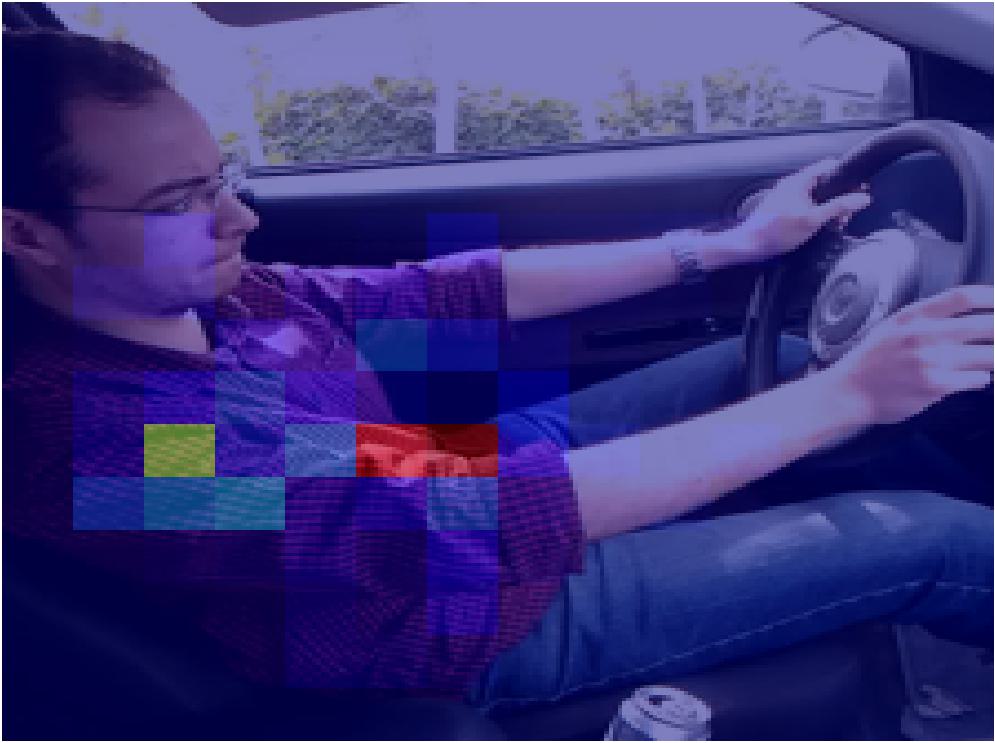} 
\includegraphics[width=\sizeafigtwo\linewidth,height=\sizeafigtwo\linewidth]{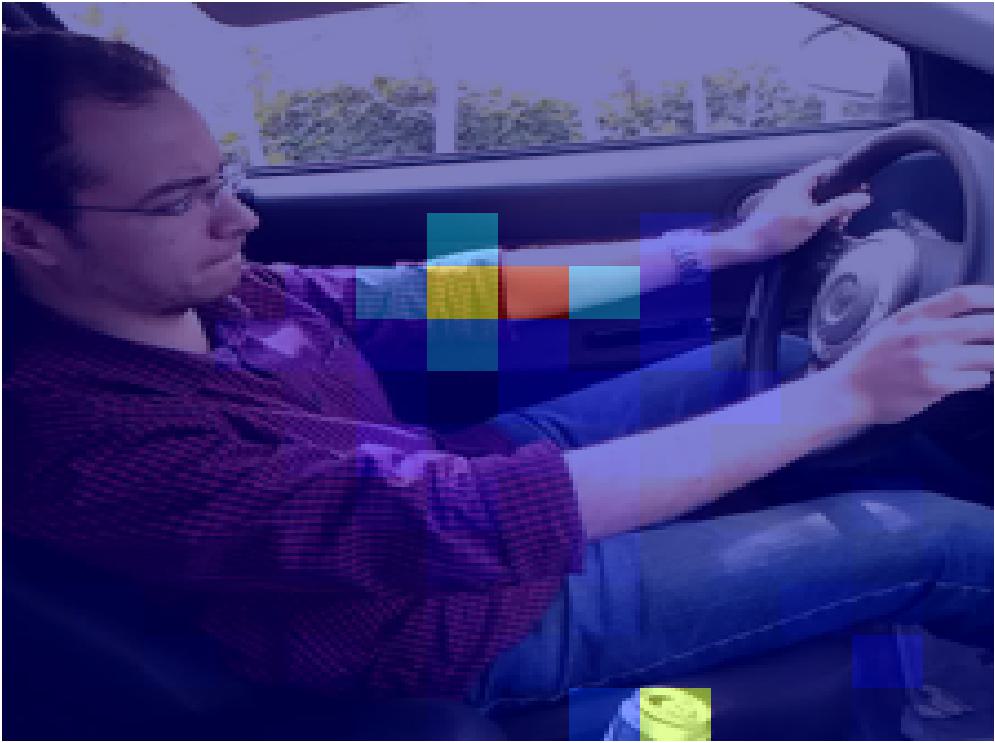}
\includegraphics[width=\sizeafigtwo\linewidth,height=\sizeafigtwo\linewidth]{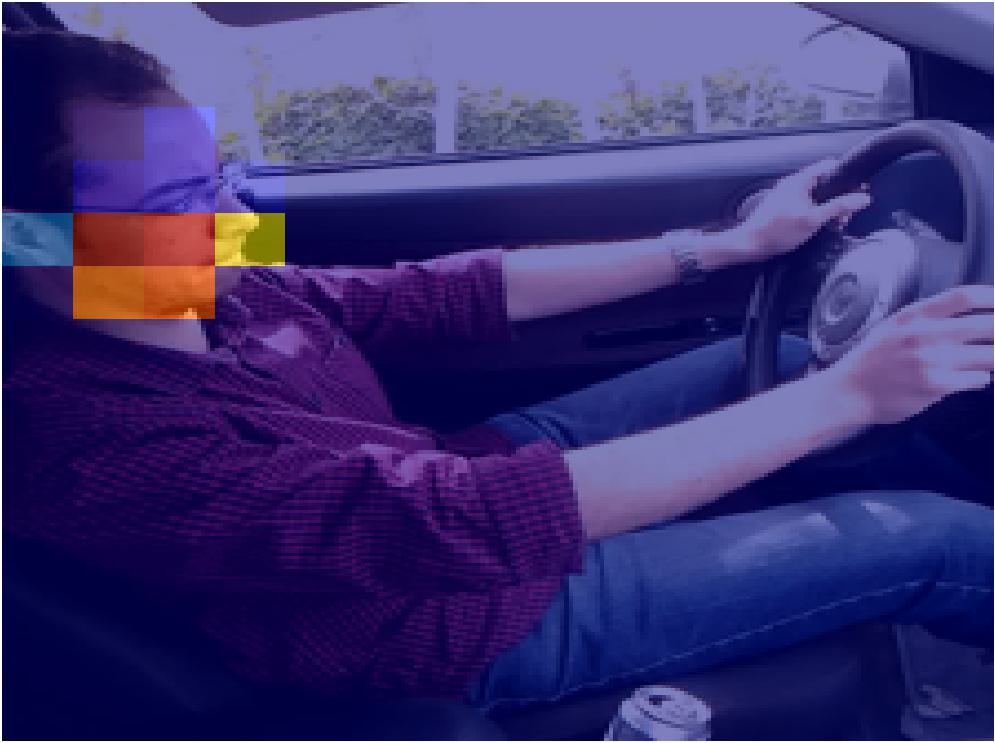}
\includegraphics[width=\sizeafigtwo\linewidth,height=\sizeafigtwo\linewidth]{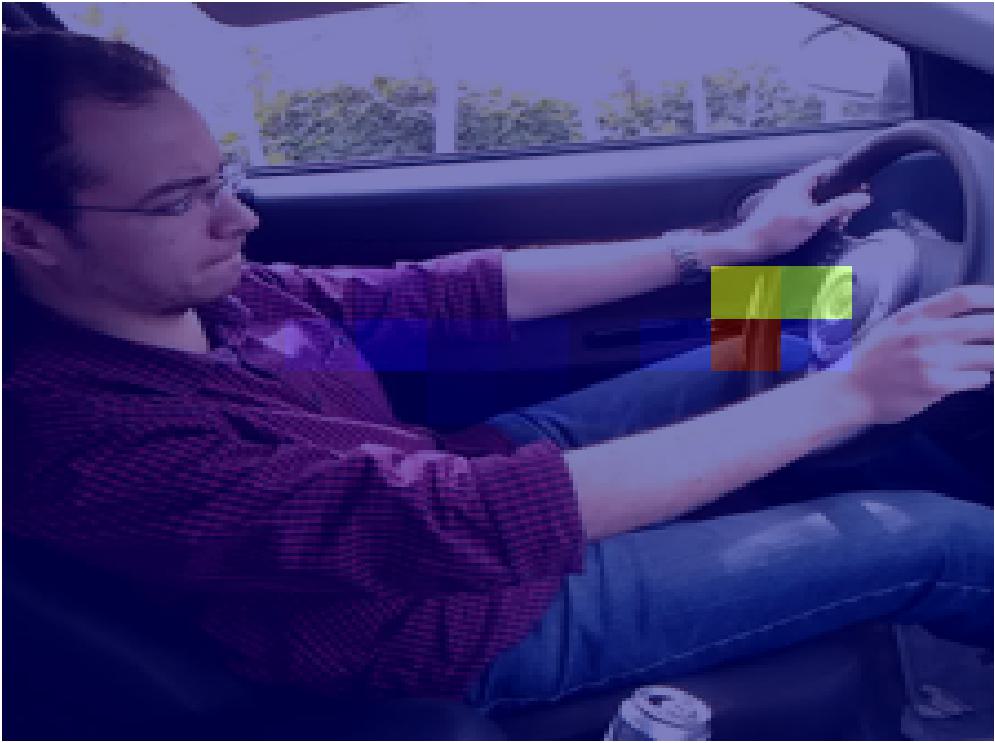}
\includegraphics[width=\sizeafigtwo\linewidth,height=\sizeafigtwo\linewidth]{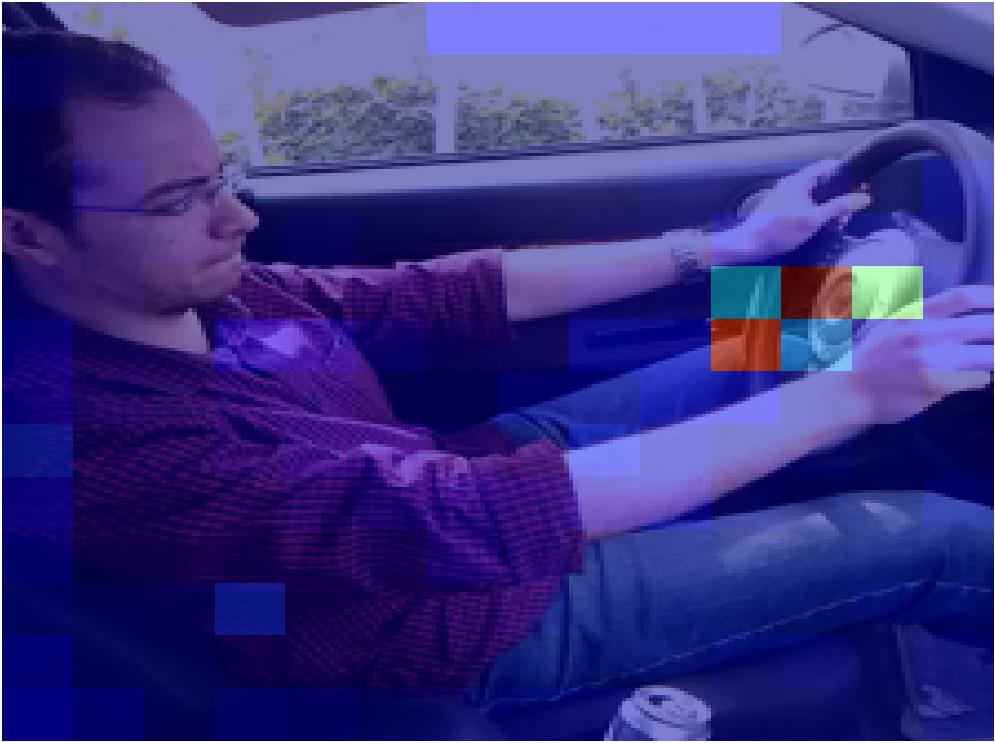}
\includegraphics[width=\sizeafigtwo\linewidth,height=\sizeafigtwo\linewidth]{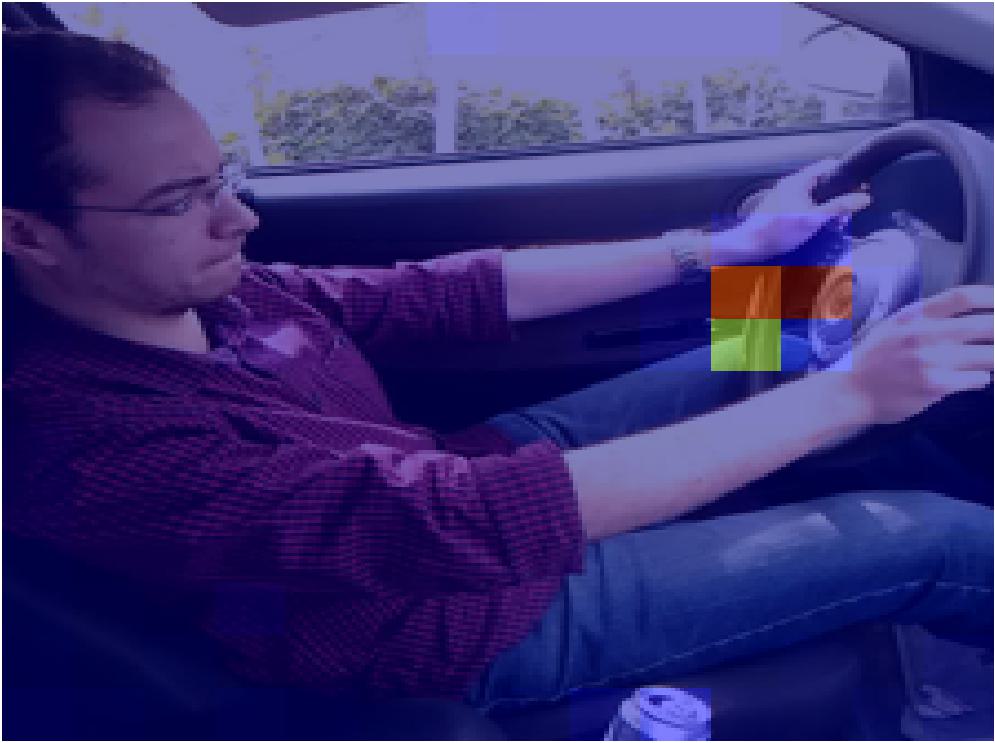} \\
Layer 7 to Layer 12
\vspacefigtwo
\end{minipage}
\end{minipage}

\hbox to \textwidth{\leaders\hbox to 6pt{\hss . \hss}\hfil}
\vspacefigtwo

\begin{minipage}{\sizelfigtwo\textwidth}
\centering
\fontsizefigtwo
\includegraphics[width=0.4\linewidth]{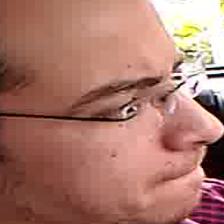}\\
Fear
\end{minipage}
\begin{minipage}{\sizerfigtwo\textwidth}
\begin{minipage}{\textwidth}
\centering
\fontsizefigtwo
\includegraphics[width=\sizebfigtwo\textwidth,height=\sizebfigtwo\textwidth]{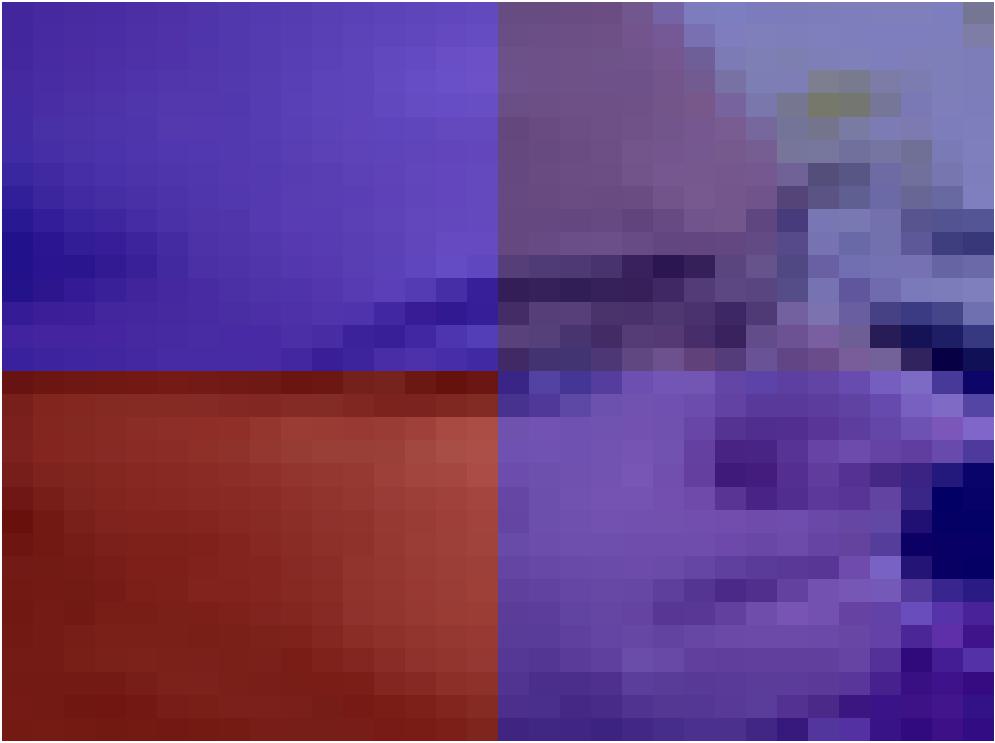}
\includegraphics[width=\sizebfigtwo\textwidth,height=\sizebfigtwo\textwidth]{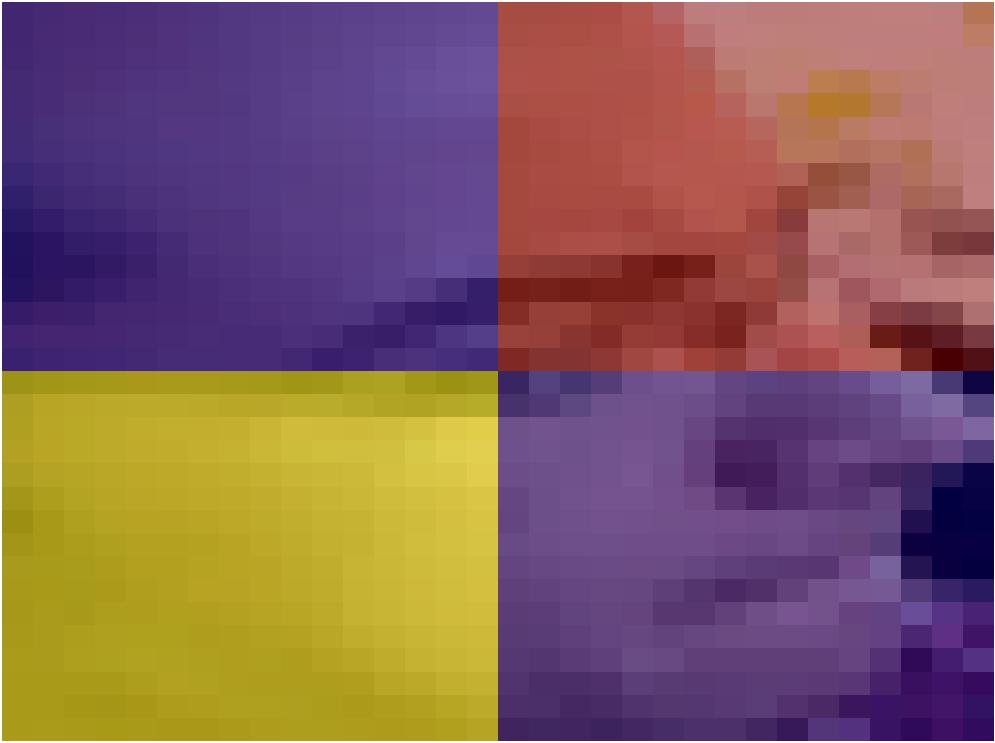}
\includegraphics[width=\sizebfigtwo\textwidth,height=\sizebfigtwo\textwidth]{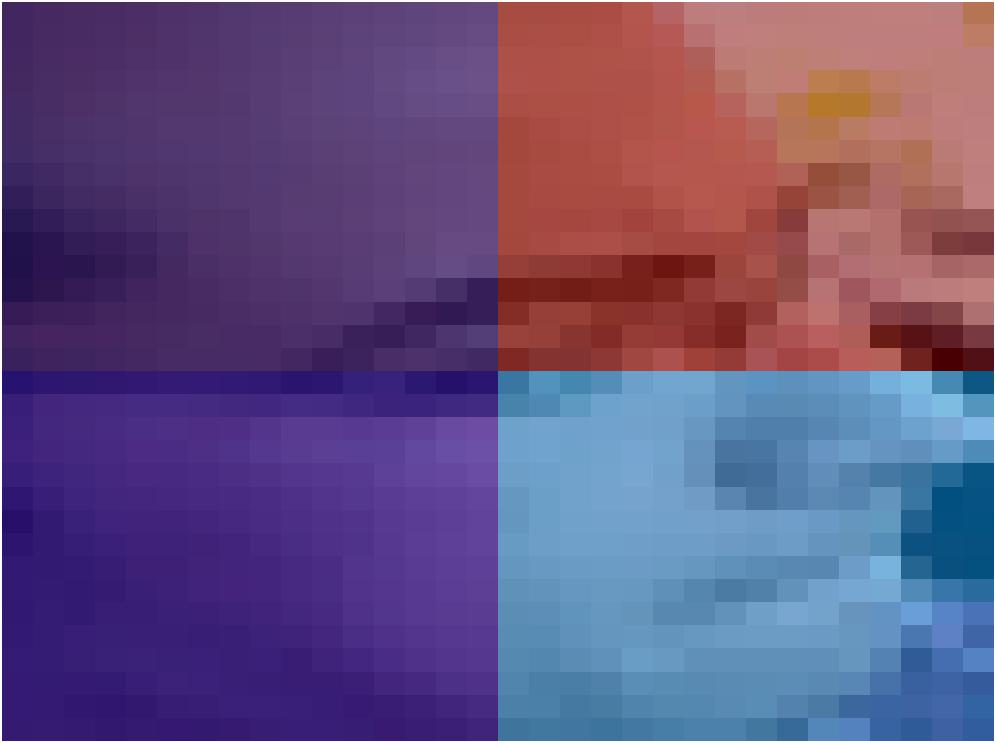}
\includegraphics[width=\sizebfigtwo\textwidth,height=\sizebfigtwo\textwidth]{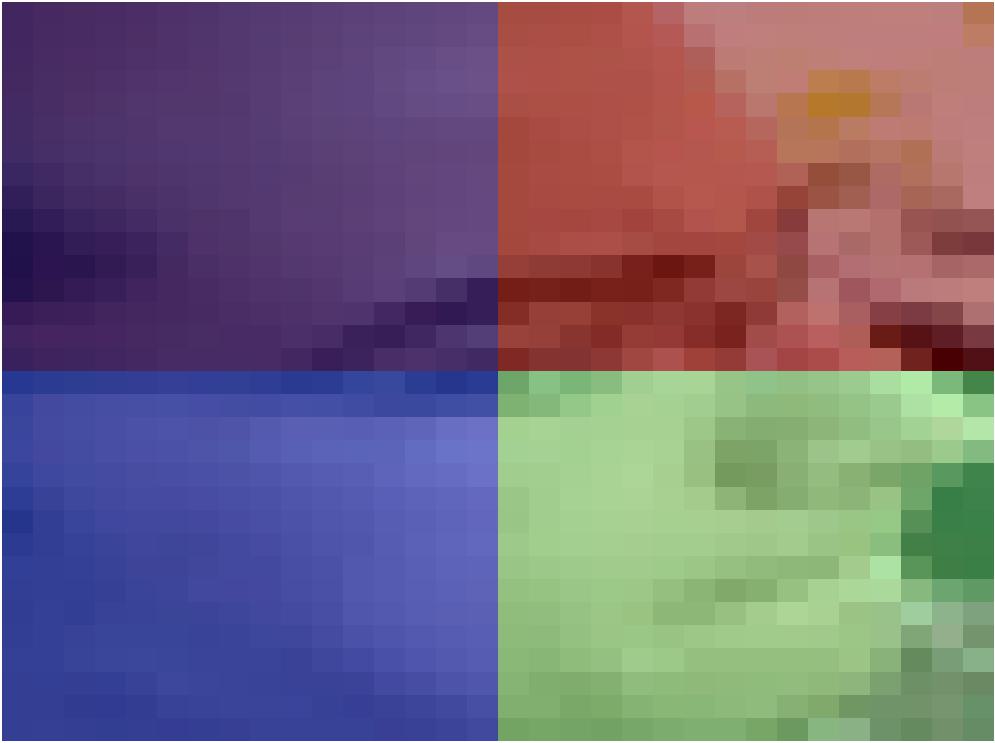}
\includegraphics[width=\sizebfigtwo\textwidth,height=\sizebfigtwo\textwidth]{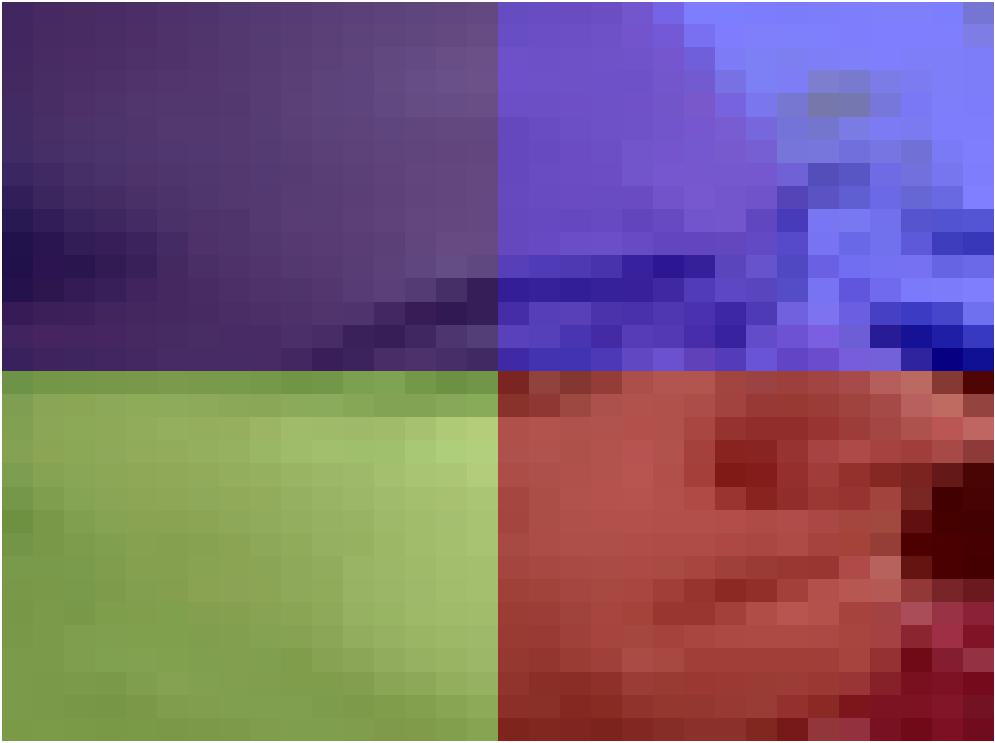}
\includegraphics[width=\sizebfigtwo\textwidth,height=\sizebfigtwo\textwidth]{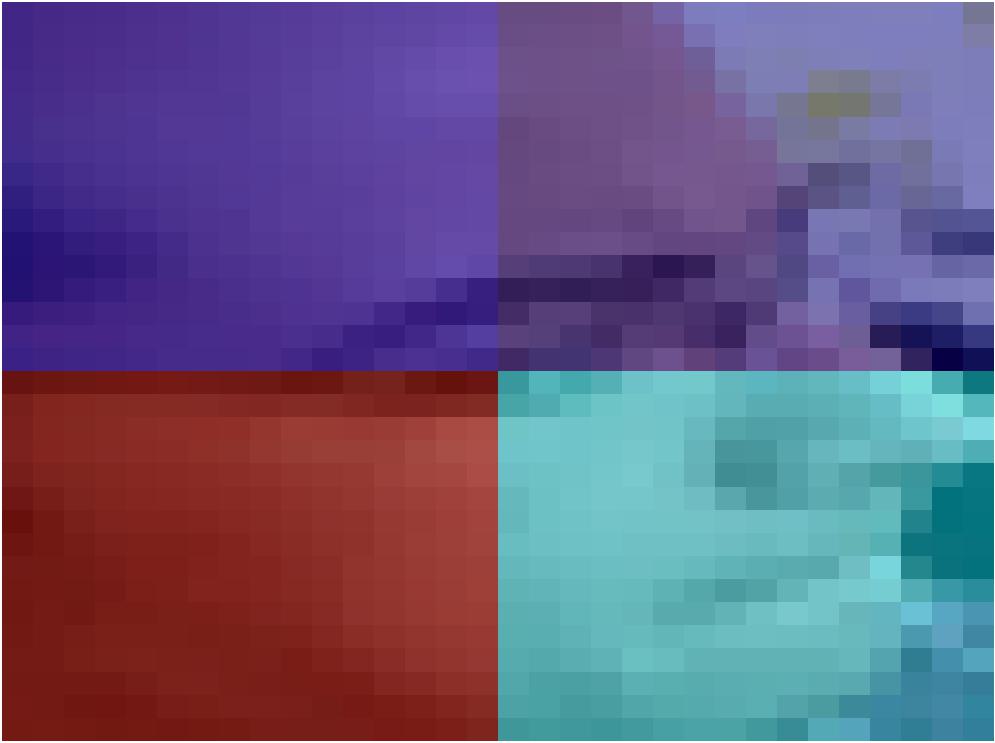}
\includegraphics[width=\sizebfigtwo\textwidth,height=\sizebfigtwo\textwidth]{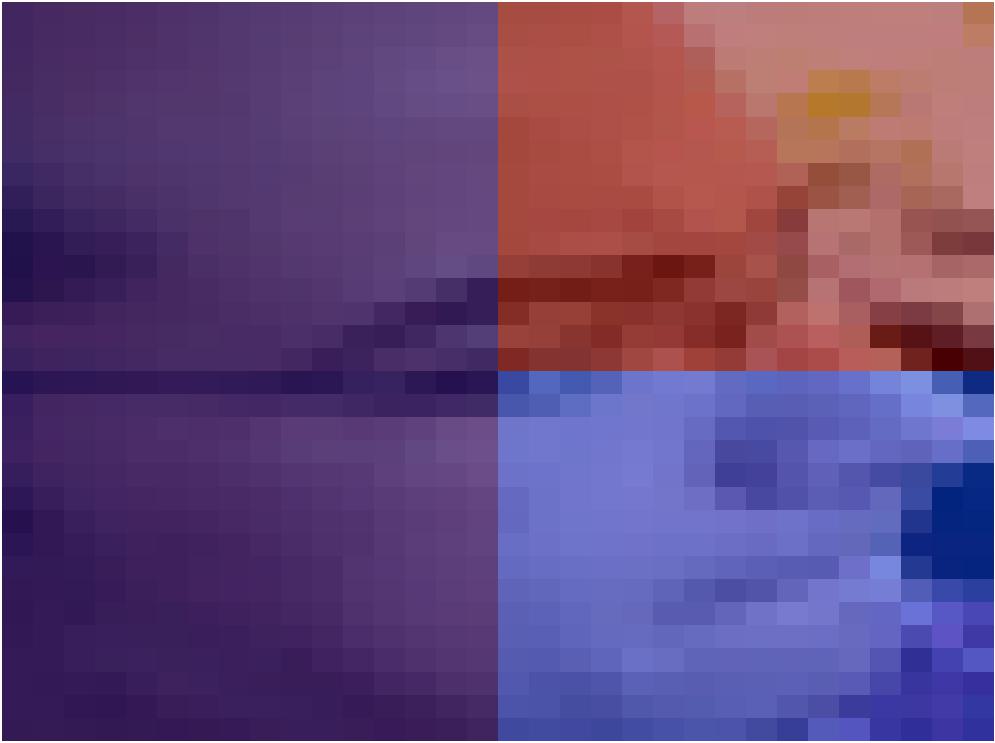}
\includegraphics[width=\sizebfigtwo\textwidth,height=\sizebfigtwo\textwidth]{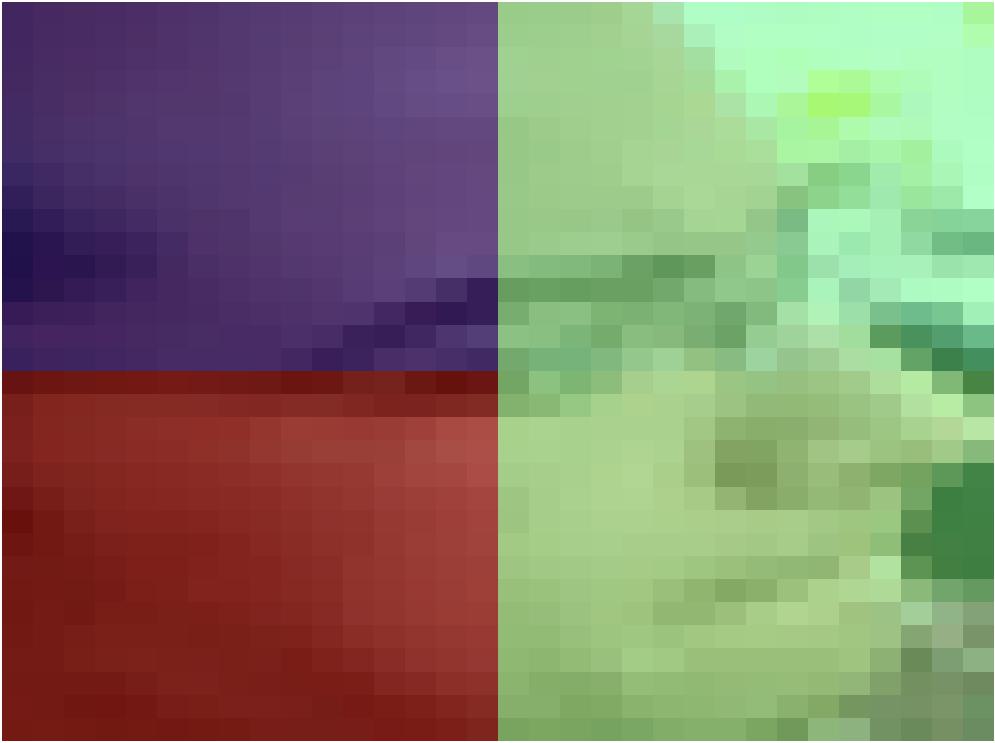}
\includegraphics[width=\sizebfigtwo\textwidth,height=\sizebfigtwo\textwidth]{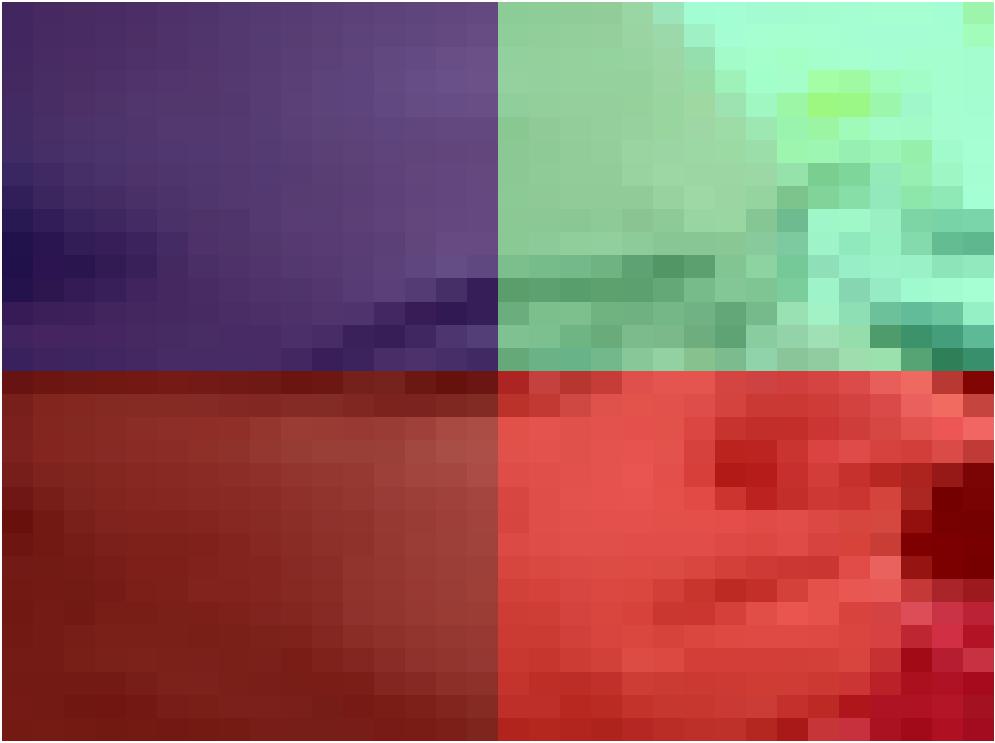}
\includegraphics[width=\sizebfigtwo\textwidth,height=\sizebfigtwo\textwidth]{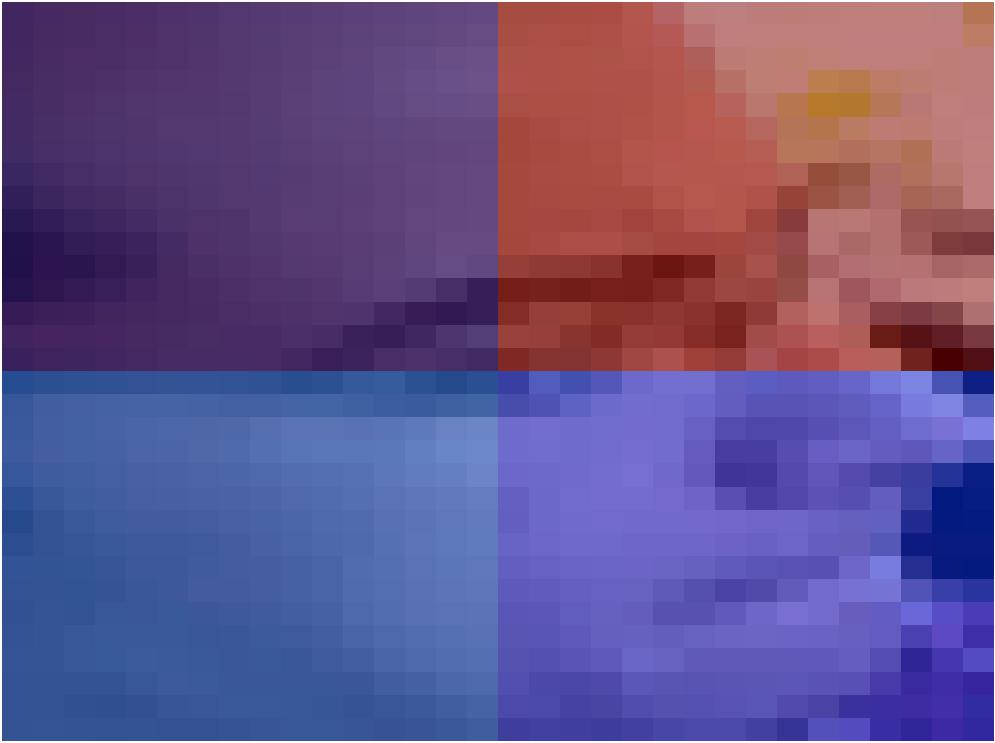}
\includegraphics[width=\sizebfigtwo\textwidth,height=\sizebfigtwo\textwidth]{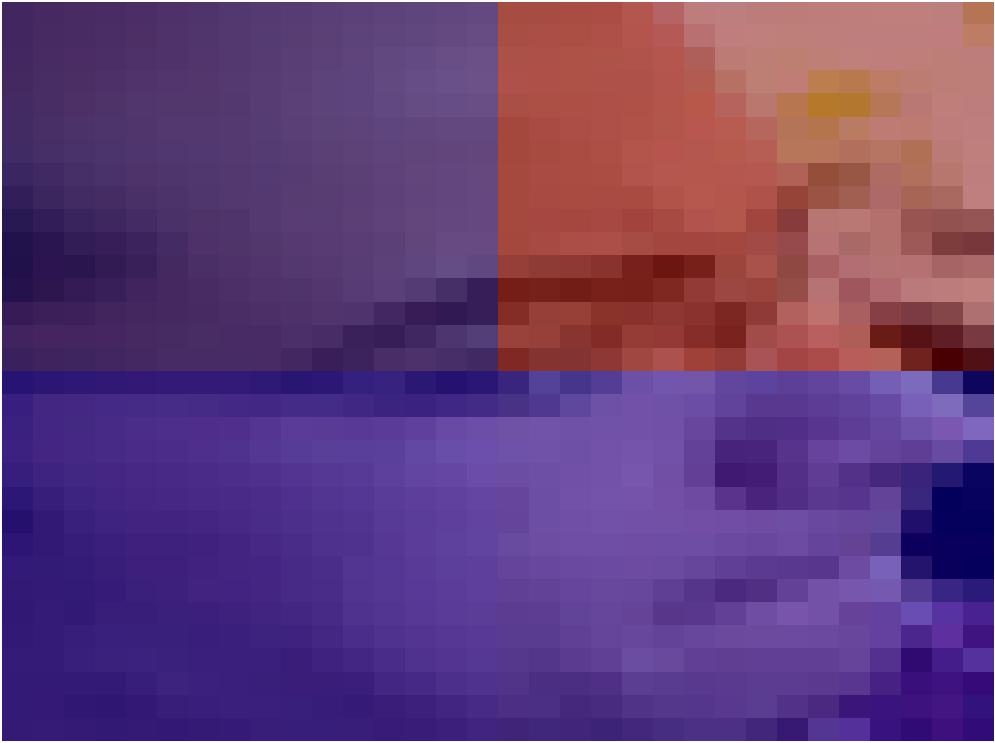}
\includegraphics[width=\sizebfigtwo\textwidth,height=\sizebfigtwo\textwidth]{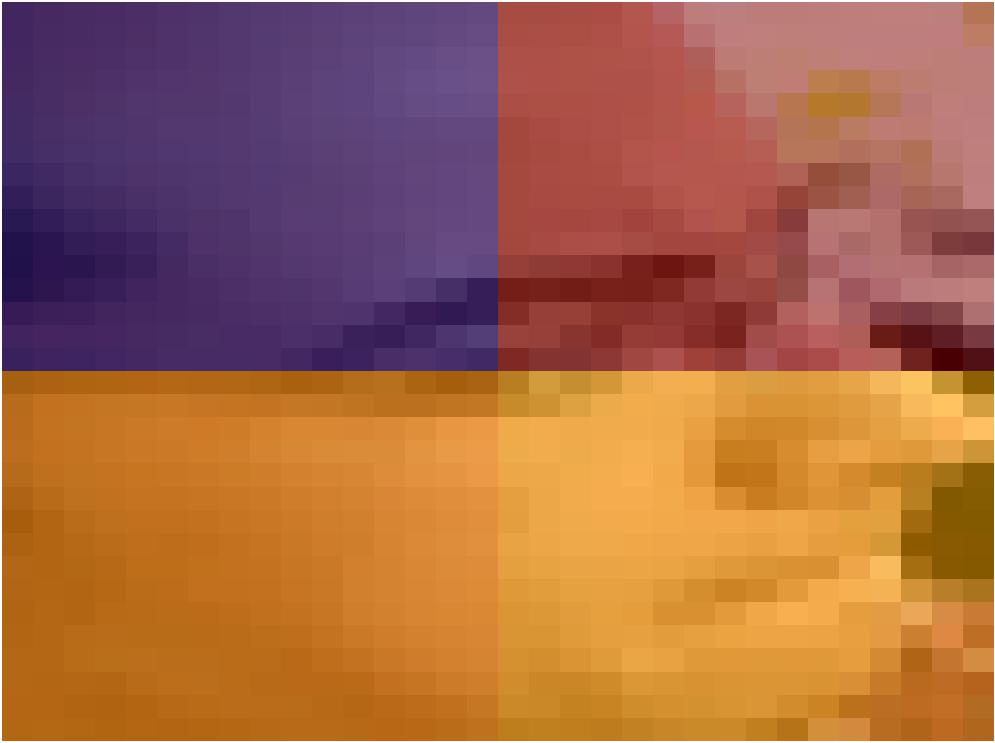} \\
Layer 1 to Layer 12
\end{minipage}
\end{minipage}

\vspacefigtwo
\hbox to \textwidth{\leaders\hbox to 6pt{\hss - \hss}\hfil}
\vspacefigtwo

\begin{minipage}{\sizelfigtwo\textwidth}
\centering
\fontsizefigtwo
\includegraphics[width=0.8\textwidth]{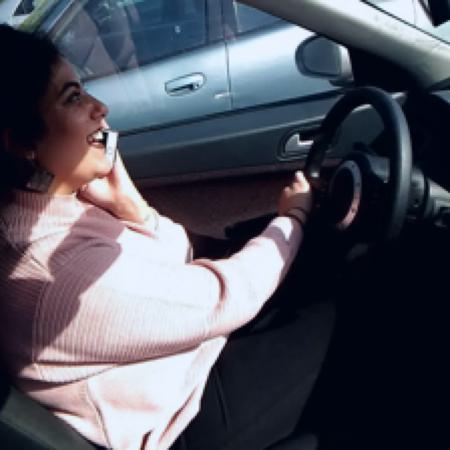}\\
Phone Left
\end{minipage}
\begin{minipage}{\sizerfigtwo\textwidth}
\centering
\begin{minipage}{\textwidth}
\centering
\fontsizefigtwo
\includegraphics[width=\sizeafigtwo\linewidth,height=\sizeafigtwo\linewidth]{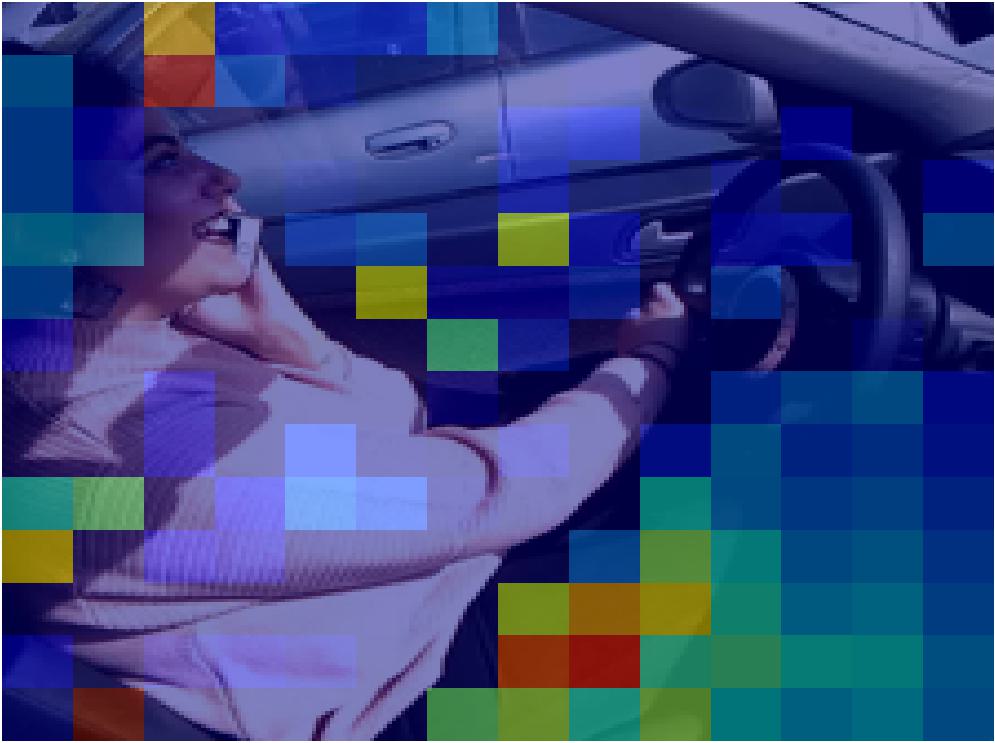}
\includegraphics[width=\sizeafigtwo\linewidth,height=\sizeafigtwo\linewidth]{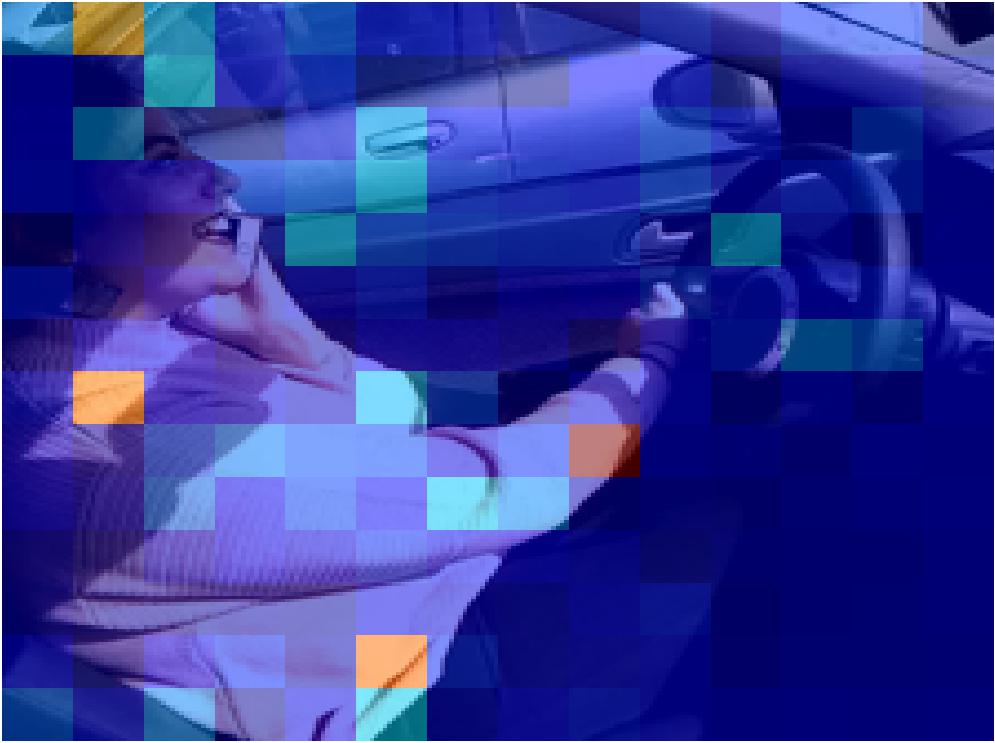}
\includegraphics[width=\sizeafigtwo\linewidth,height=\sizeafigtwo\linewidth]{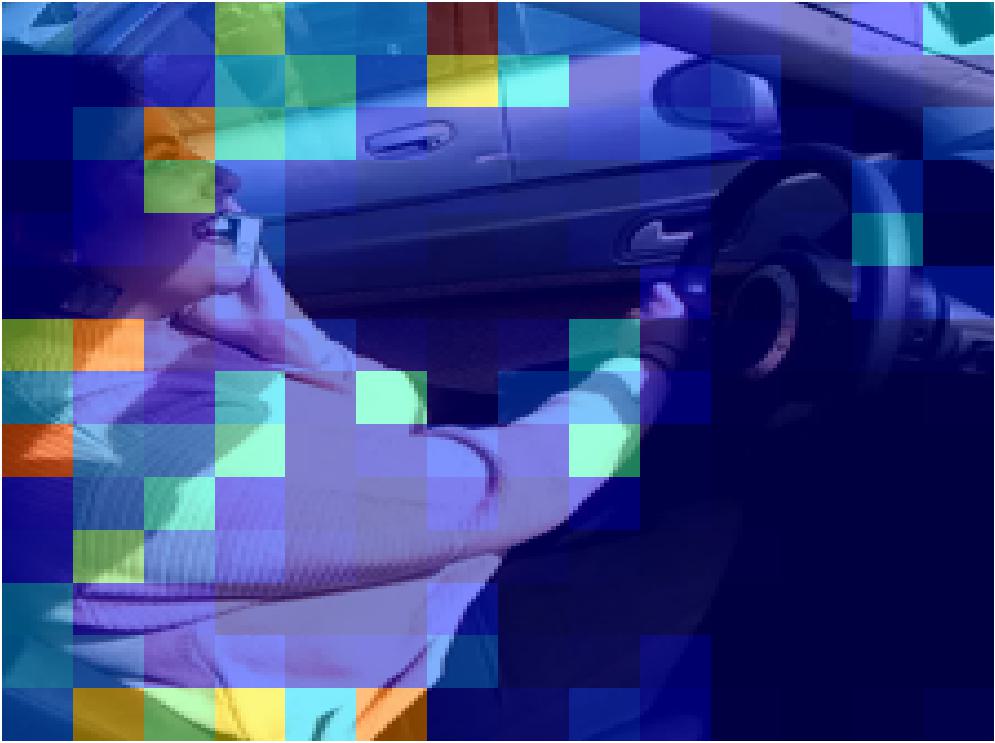}
\includegraphics[width=\sizeafigtwo\linewidth,height=\sizeafigtwo\linewidth]{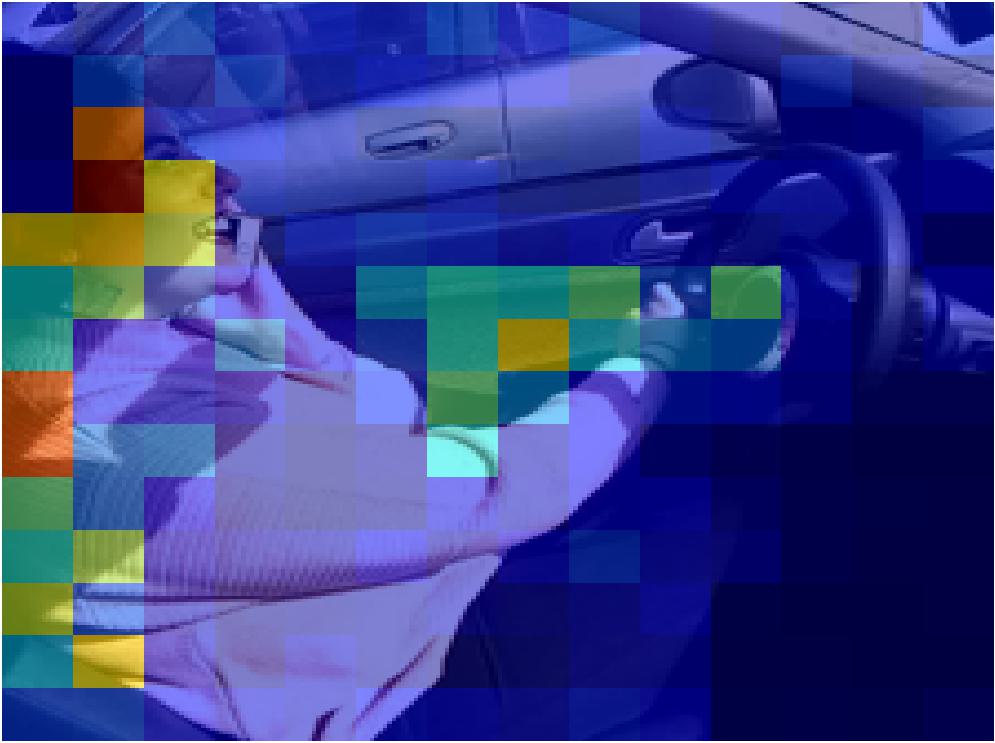}
\includegraphics[width=\sizeafigtwo\linewidth,height=\sizeafigtwo\linewidth]{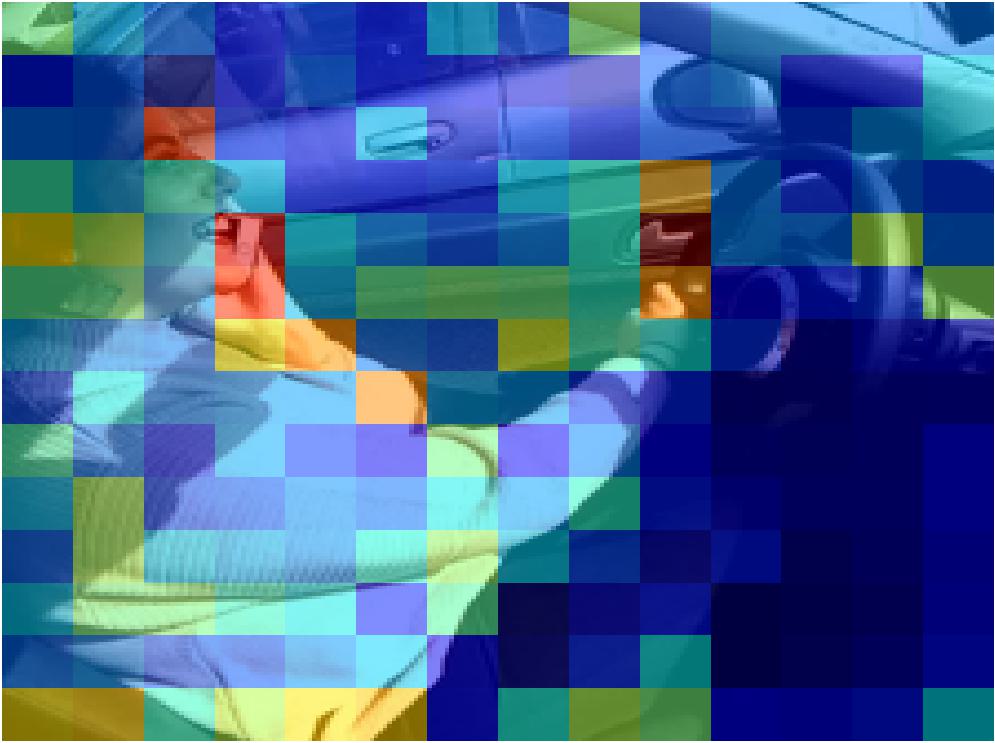} 
\includegraphics[width=\sizeafigtwo\linewidth,height=\sizeafigtwo\linewidth]{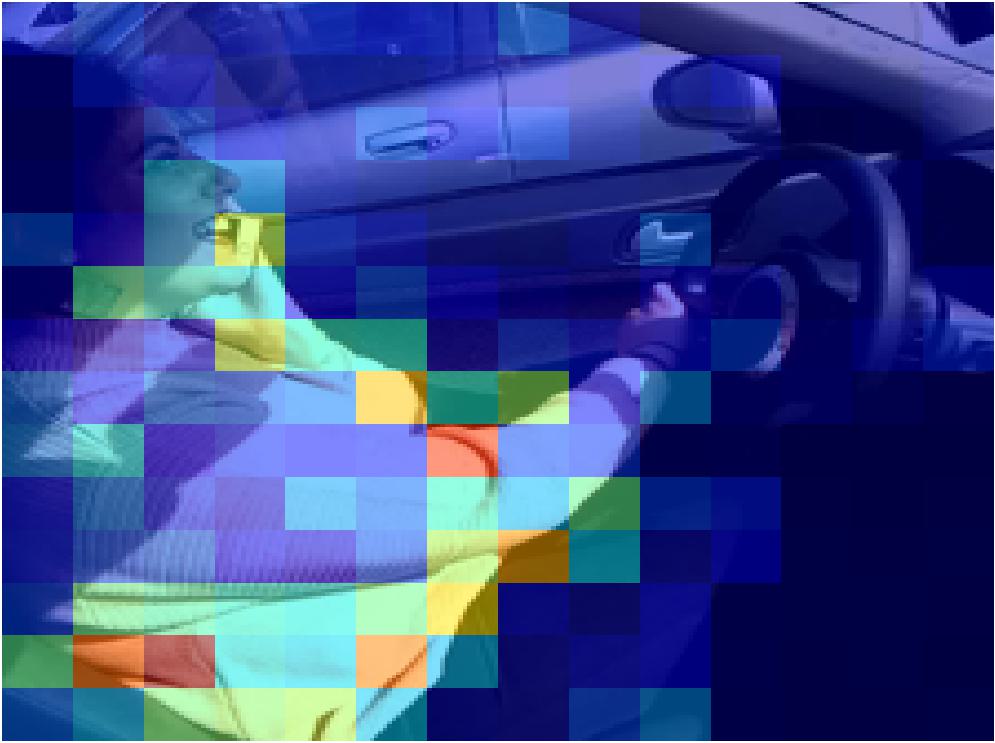} \\
Layer 1 to Layer 6
\vspacefigtwo
\end{minipage}
\begin{minipage}{\textwidth}
\centering
\fontsizefigtwo
\includegraphics[width=\sizeafigtwo\linewidth,height=\sizeafigtwo\linewidth]{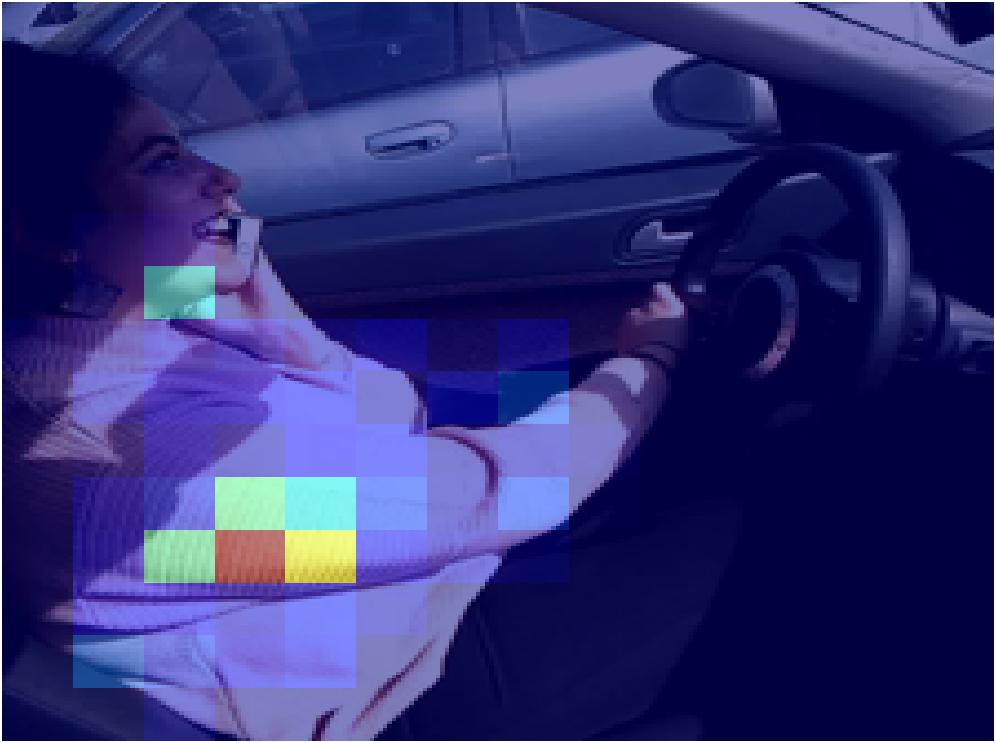} 
\includegraphics[width=\sizeafigtwo\linewidth,height=\sizeafigtwo\linewidth]{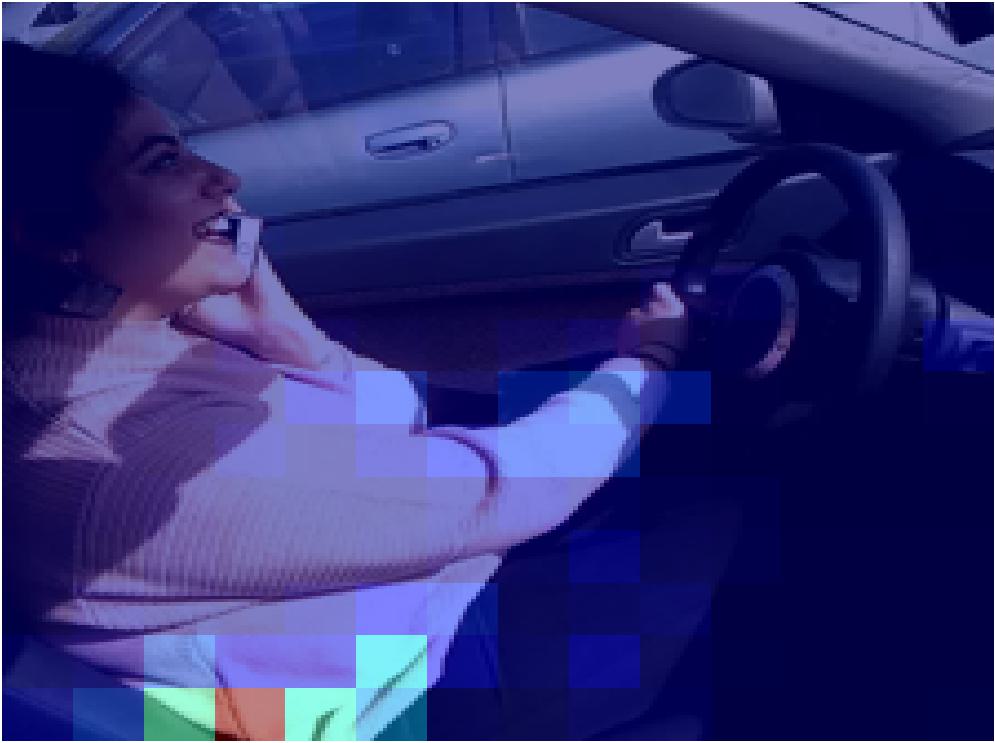}
\includegraphics[width=\sizeafigtwo\linewidth,height=\sizeafigtwo\linewidth]{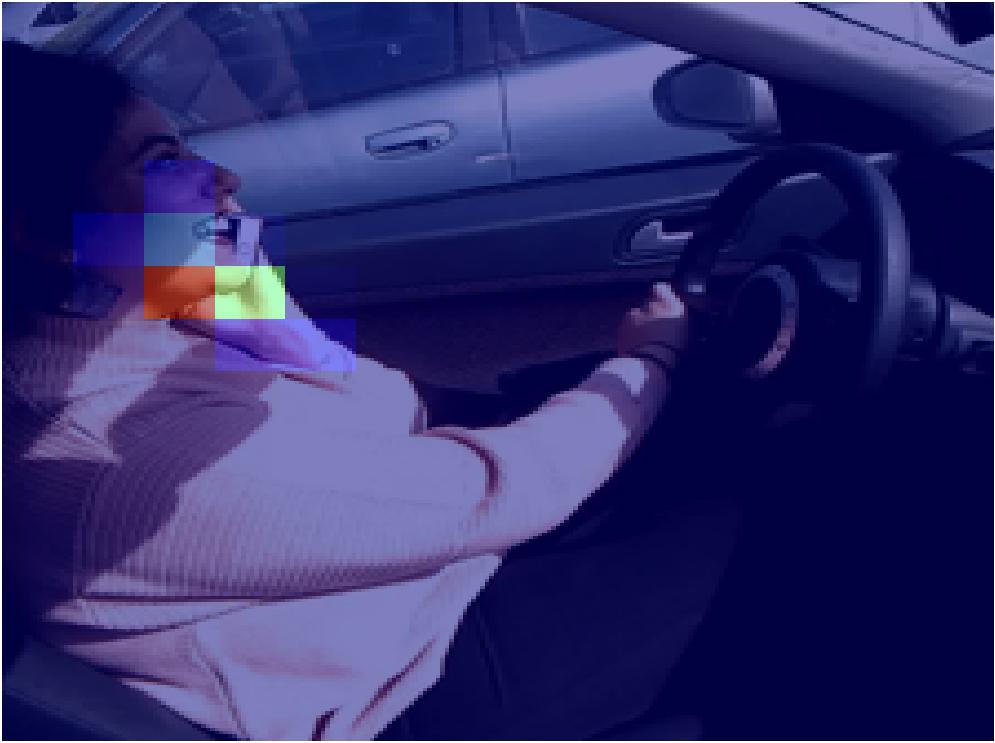}
\includegraphics[width=\sizeafigtwo\linewidth,height=\sizeafigtwo\linewidth]{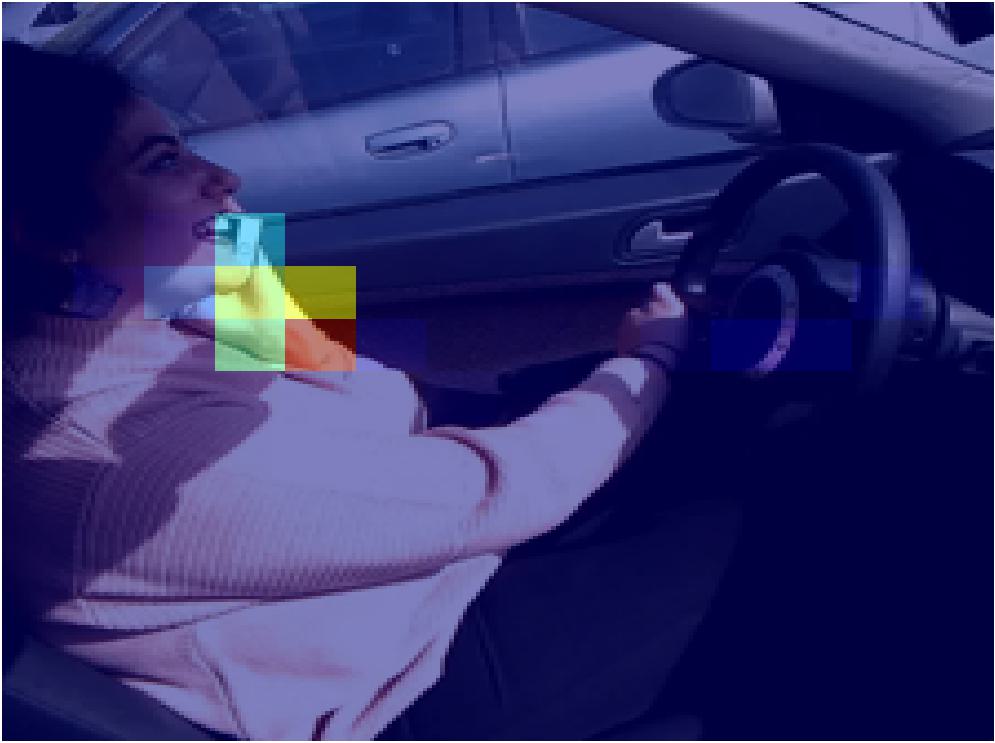}
\includegraphics[width=\sizeafigtwo\linewidth,height=\sizeafigtwo\linewidth]{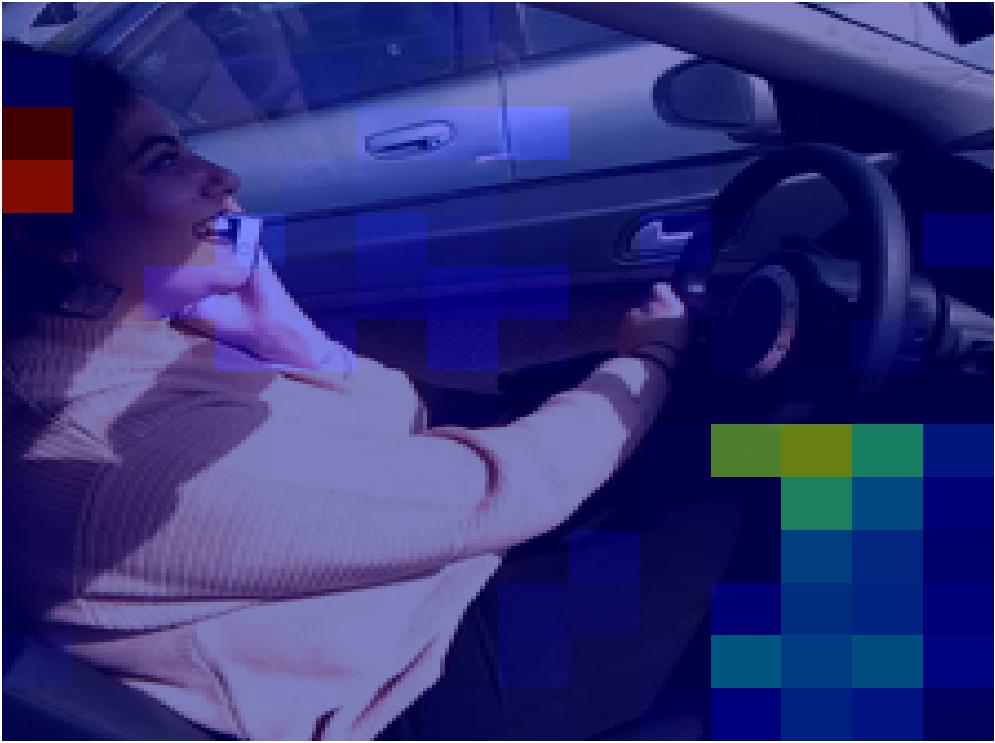}
\includegraphics[width=\sizeafigtwo\linewidth,height=\sizeafigtwo\linewidth]{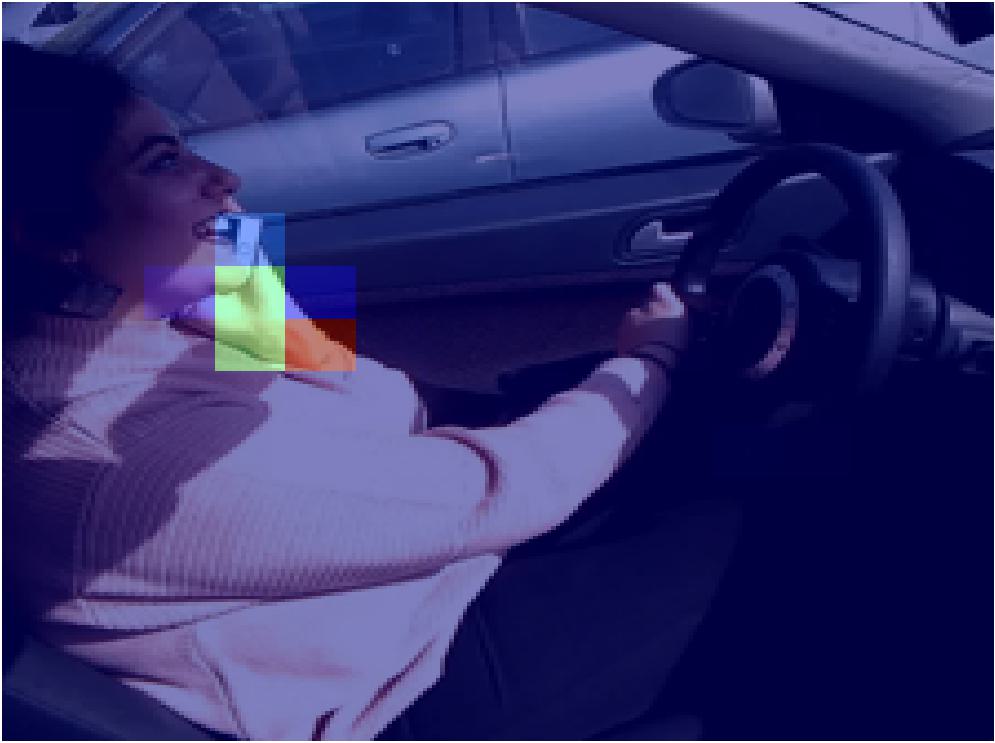} \\
Layer 7 to Layer 12
\vspacefigtwo
\end{minipage}
\end{minipage}

\hbox to \textwidth{\leaders\hbox to 6pt{\hss . \hss}\hfil}
\vspacefigtwo

\begin{minipage}{\sizelfigtwo\textwidth}
\centering
\fontsizefigtwo
\includegraphics[width=0.4\textwidth]{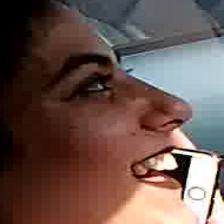}\\
Happy
\end{minipage}
\begin{minipage}{\sizerfigtwo\textwidth}
\begin{minipage}{\textwidth}
\centering
\fontsizefigtwo
\includegraphics[width=\sizebfigtwo\linewidth,height=\sizebfigtwo\linewidth]{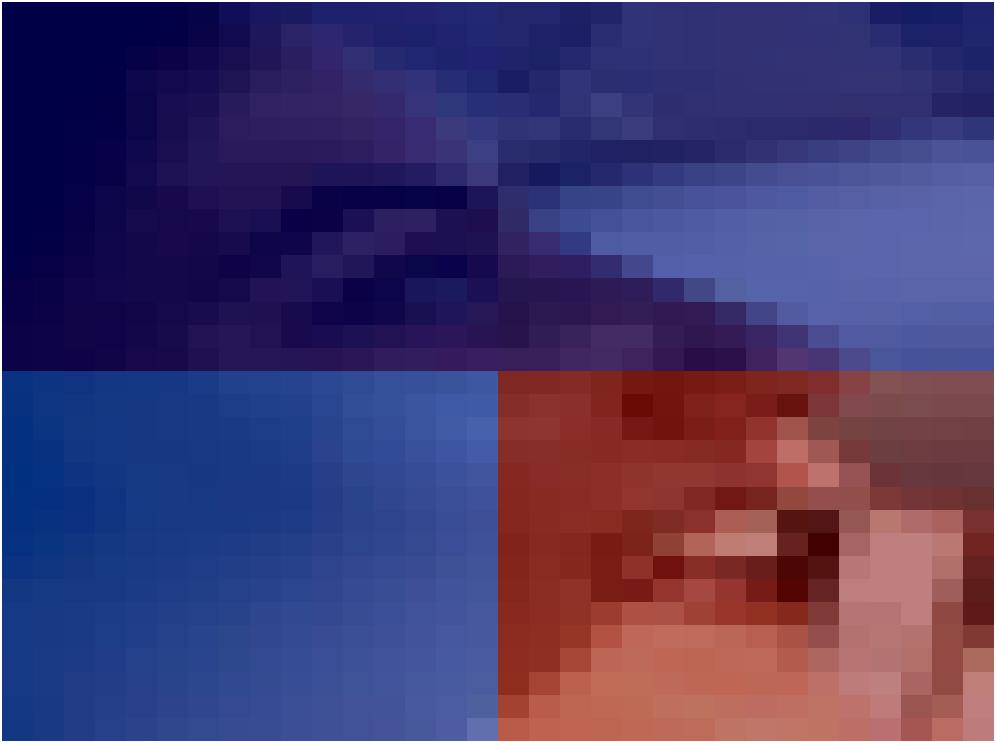}
\includegraphics[width=\sizebfigtwo\linewidth,height=\sizebfigtwo\linewidth]{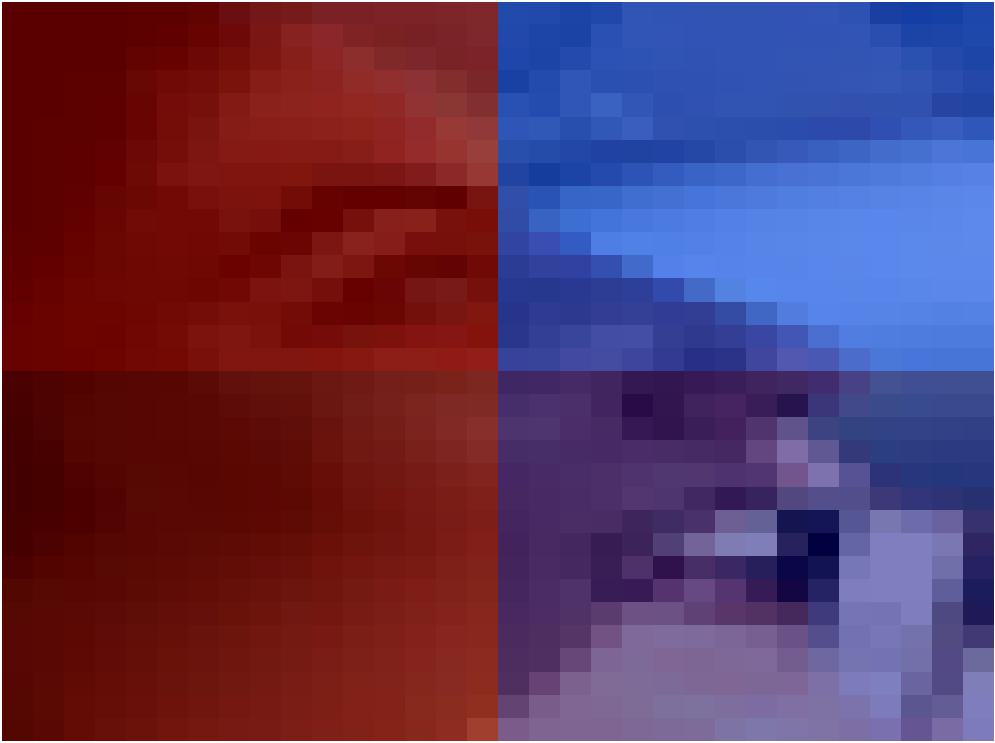}
\includegraphics[width=\sizebfigtwo\linewidth,height=\sizebfigtwo\linewidth]{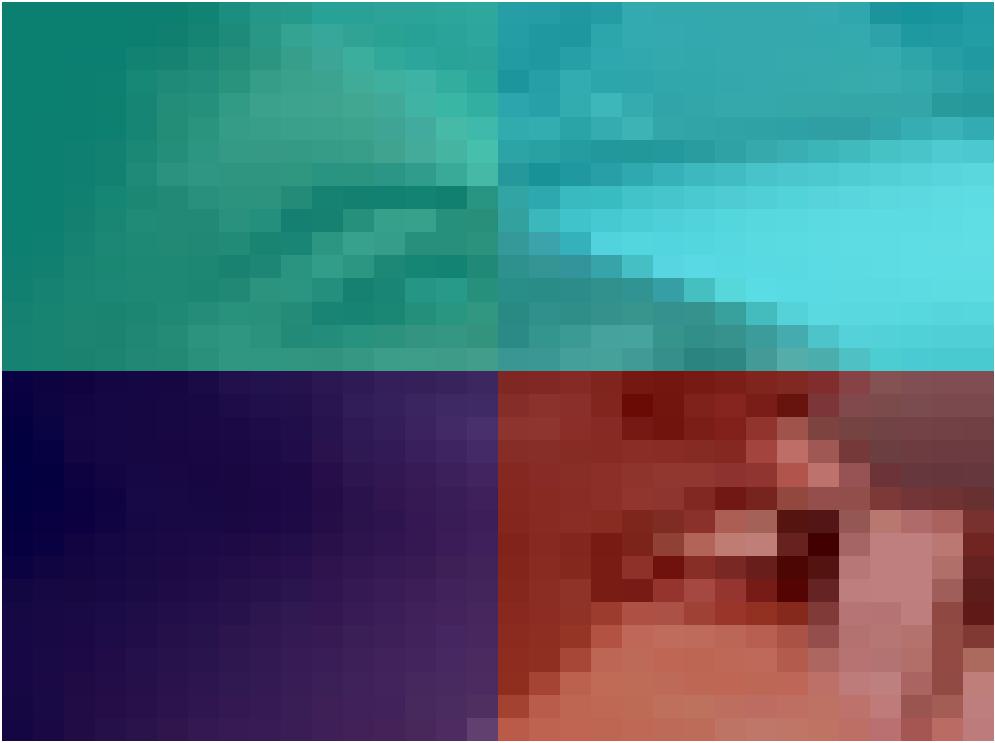}
\includegraphics[width=\sizebfigtwo\linewidth,height=\sizebfigtwo\linewidth]{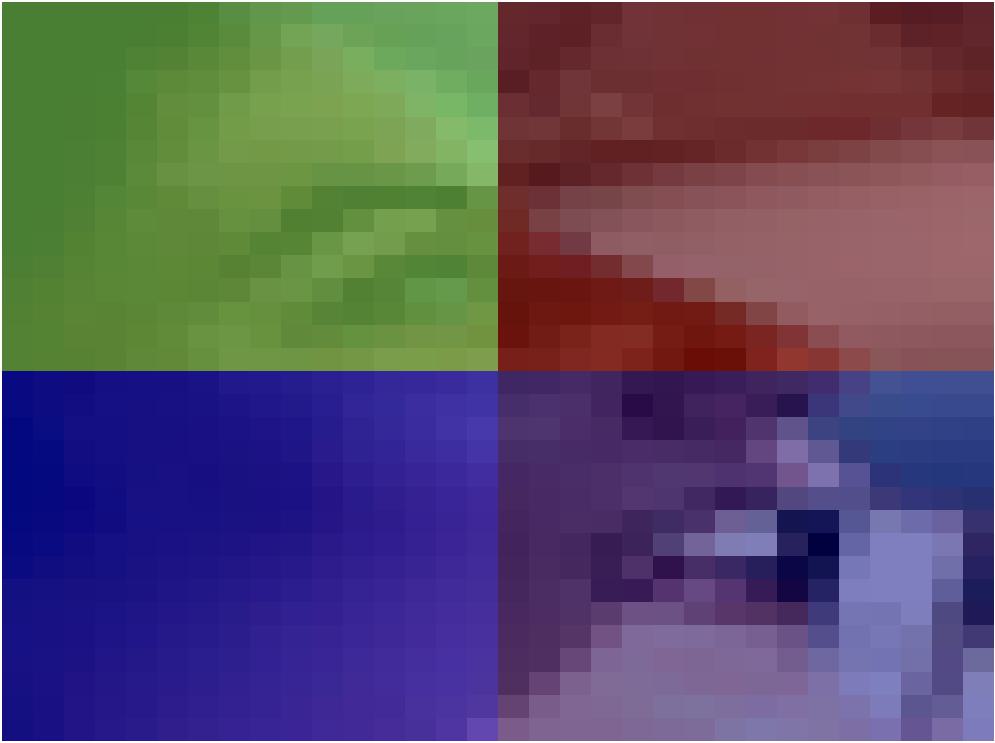}
\includegraphics[width=\sizebfigtwo\linewidth,height=\sizebfigtwo\linewidth]{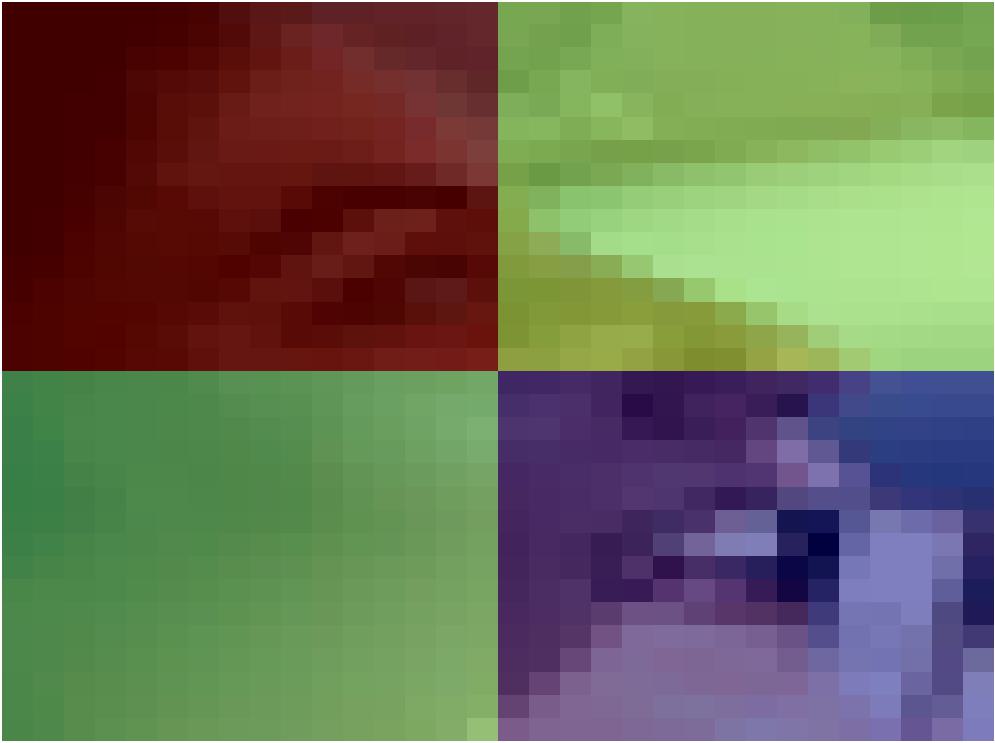}
\includegraphics[width=\sizebfigtwo\linewidth,height=\sizebfigtwo\linewidth]{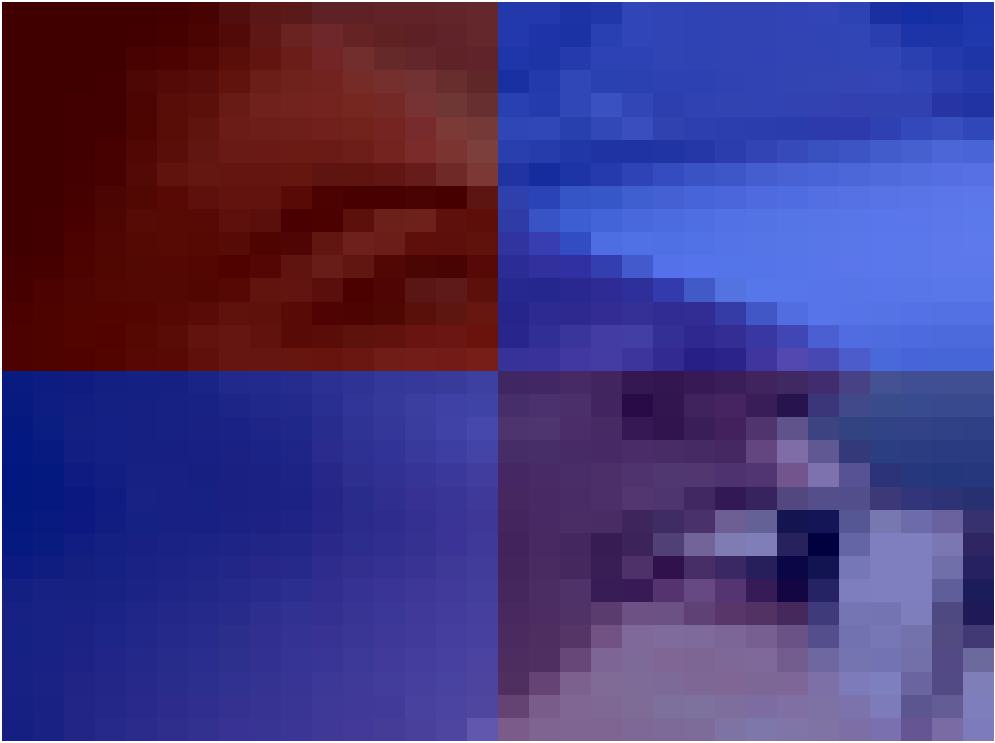}
\includegraphics[width=\sizebfigtwo\linewidth,height=\sizebfigtwo\linewidth]{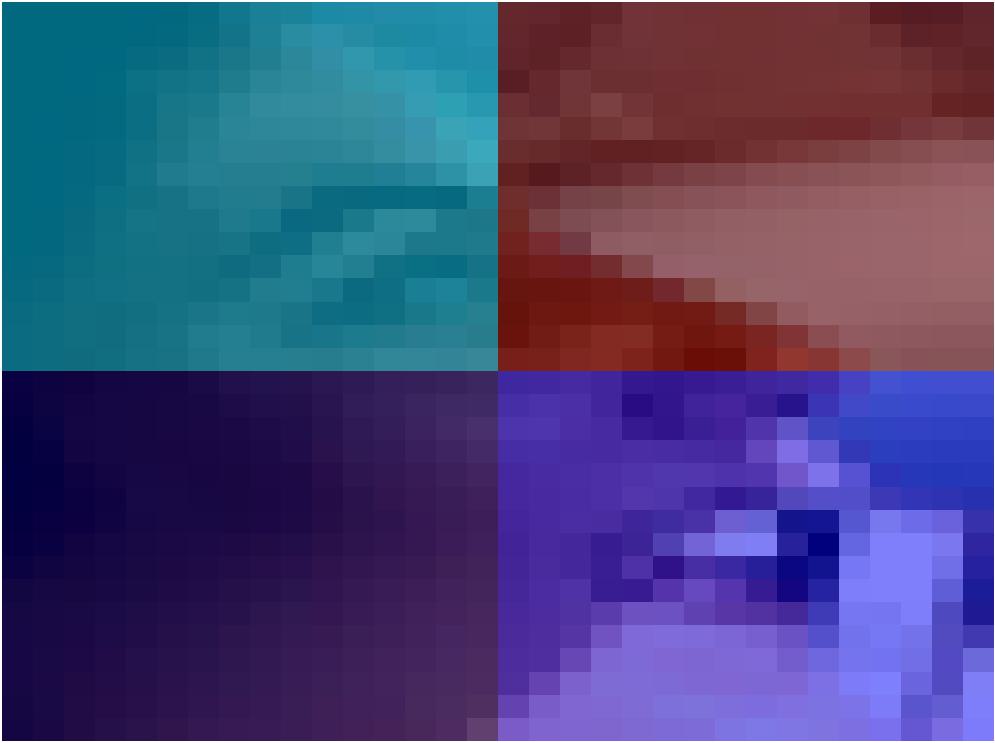}
\includegraphics[width=\sizebfigtwo\linewidth,height=\sizebfigtwo\linewidth]{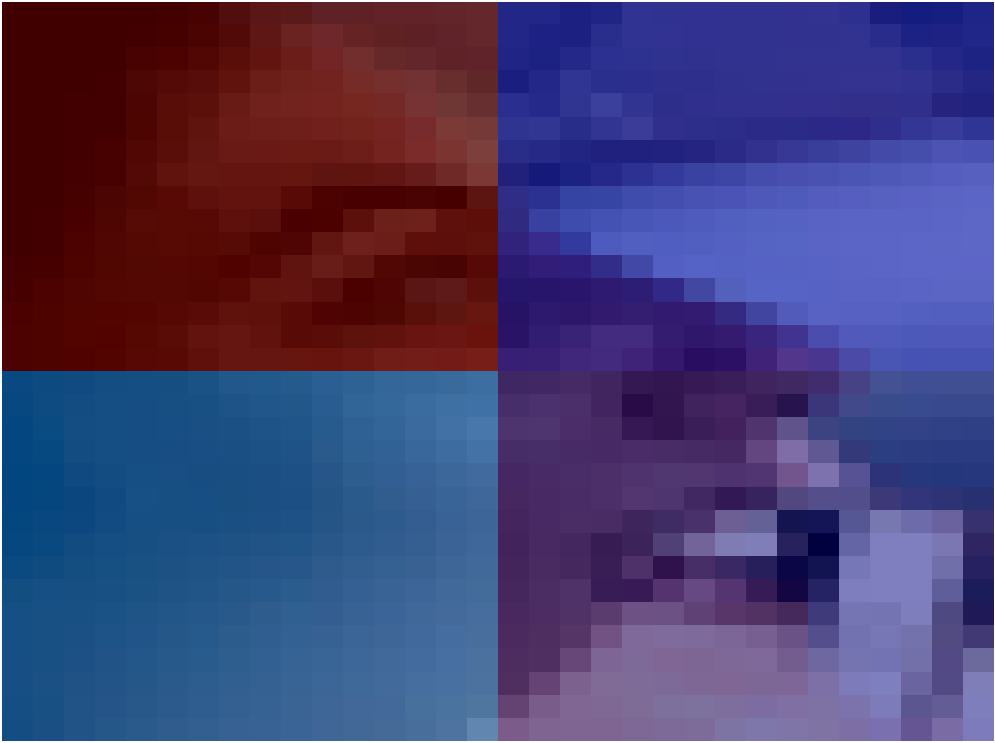}
\includegraphics[width=\sizebfigtwo\linewidth,height=\sizebfigtwo\linewidth]{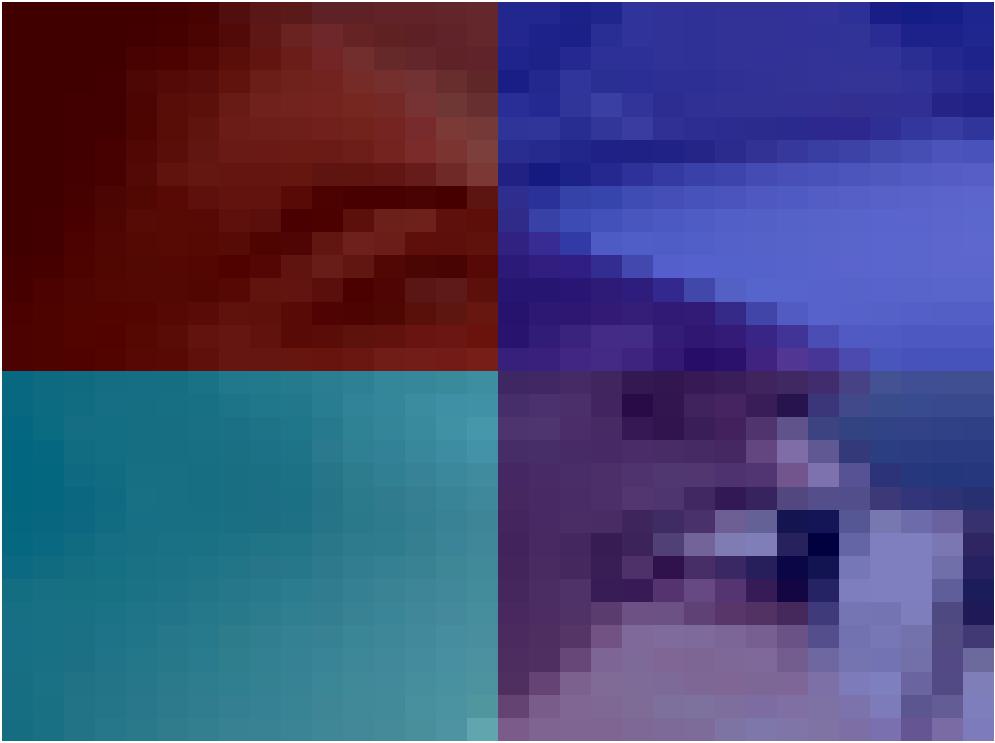}
\includegraphics[width=\sizebfigtwo\linewidth,height=\sizebfigtwo\linewidth]{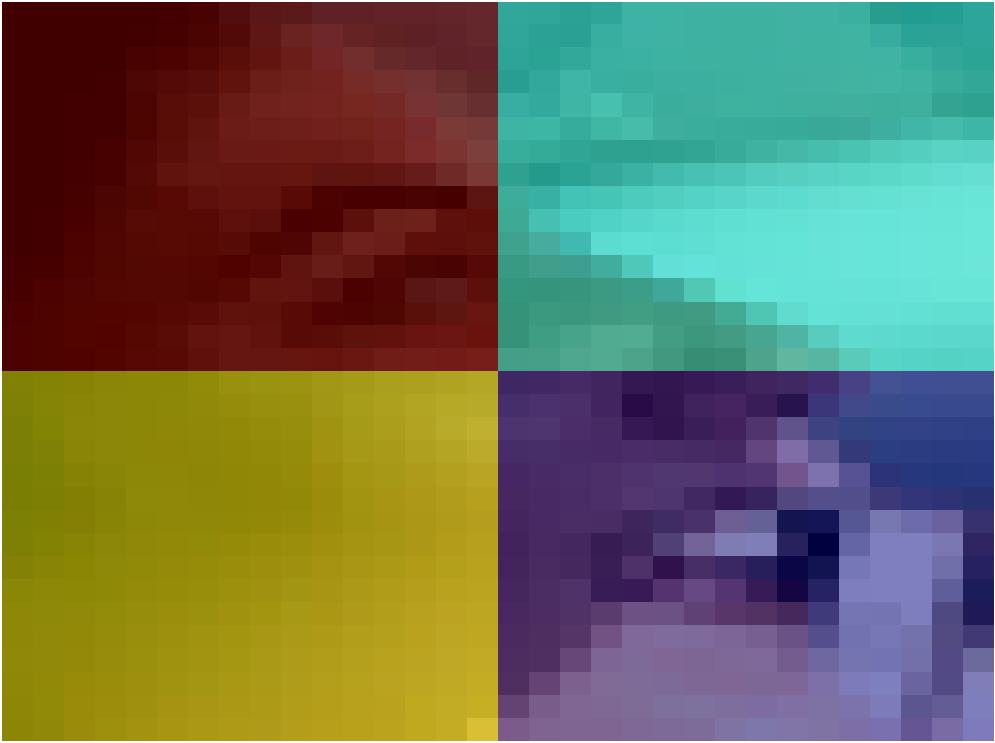}
\includegraphics[width=\sizebfigtwo\linewidth,height=\sizebfigtwo\linewidth]{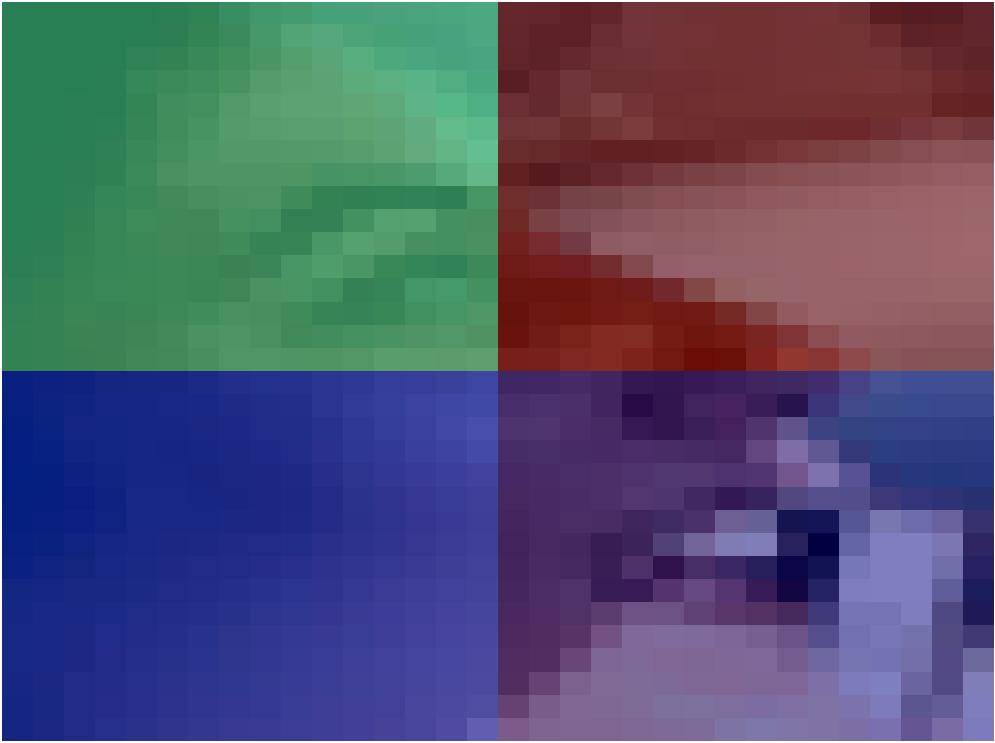}
\includegraphics[width=\sizebfigtwo\linewidth,height=\sizebfigtwo\linewidth]{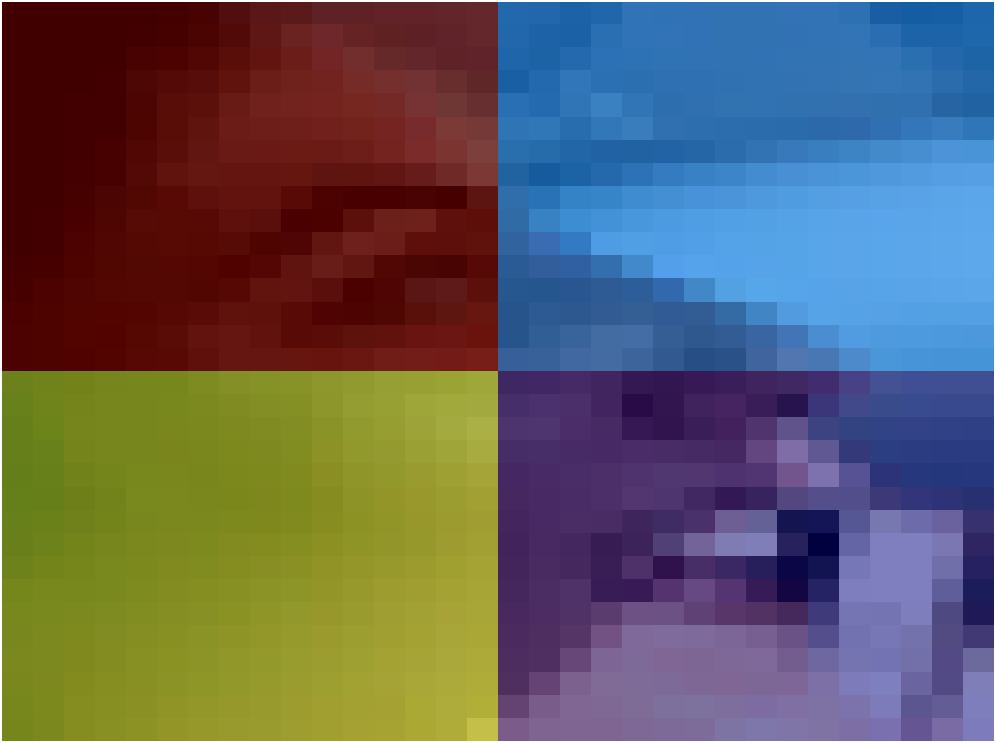}\\
Layer 1 to Layer 12
\end{minipage}
\end{minipage}
\vspacefigtwo
\end{minipage}
\caption{This figure displays the attention maps generated by a well-trained ViT-DD model during inference on the AUCDD dataset. The attention maps depict how the distraction token interacts with other visual tokens across the $L=12$ Transformer layers of ViT-DD. The colors used in the maps correspond to the level of attention, with \textcolor{red}{red} indicating \textcolor{red}{high} attention and \textcolor{blue}{blue} indicating \textcolor{blue}{low} attention. The maps show that as the network grows deeper, the distraction token focuses on precise local cues rather than the entire image signal. The model successfully concentrates on critical areas of the input images, such as the driving wheel region in the first scenario and the phone region in the second scenario.}

\label{fig:attn_map}
\end{figure*}

\section{Experiments and Results} \label{sec:Test}
In this section, the performance of the proposed ViT-DD model in detecting driver distractions is assessed. The employed benchmarks and baselines are described first. The major results are then reported along with some empirical analysis.

\subsection{Benchmarks}
The performance of ViT-DD is evaluated on two publicly available distracted driver detection datasets: State Farm Distracted Driver Detection (SFDDD) and the American University in Cairo Distracted Driver Dataset (AUCDD). The SFDDD dataset is comprised of 22,424 labeled images of 26 drivers captured by a constant-placed 2D dashboard camera with $640\times 480$ pixels in RGB \cite{sfddd}. The AUCDD dataset contains 10,555 training images and 1123 testing images with resolution of $1920\times 1080$ pixels in RGB. It includes data for 44 drivers, 38 of whom are included in the training set and 6 in the test set \cite{eraqi2019driver,abouelnaga2017real}.
Both datasets cover the same ten real-world driving postures: (C0) \textit{Safe Driving}, (C1) \textit{Phone Right}, (C2) \textit{Phone Left}, (C3) \textit{Text Right}, (C4) \textit{Text Left}, (C5) \textit{Adjusting Radio}, (C6) \textit{Drinking}, (C7) \textit{Hair or Makeup}, (C8) \textit{Reaching Behind}, and (C9) \textit{Talking to Passenger}.

There are two commonly used train/test split methods for the SFDDD dataset. One option is to directly split the dataset by images for training or testing~\cite{li2022distracted,alotaibi2020distracted,lin2022lightweight,9661254}, but it will result in a strong correlation between the training and testing data. In particular, it is possible that consecutive video frames are divided into training and testing sets, so it simplifies the problem of distraction detection. The other one is to divide the dataset by drivers such that images of the same driver do not appear in both training and test data~\cite{lu2020rcnn}. 

In this paper, results of both split approaches for the SFDDD dataset are reported. For the first split method, $70\%/30\%$ of images are used for training and testing, respectively. While for the second, 18/6 of the total 28 drivers are randomly selected for training/testing. The AUCDD dataset adheres to the original split-by-driver setup.

\subsection{Baselines}
To compare ViT-DD with the state-of-the-art approaches for distracted driver detection, the following methods are selected as baselines: (1) GA-Weighted Ensemble\cite{eraqi2019driver}, (2) ADNet\cite{xiao2022attention}, (3) C-SLSTM\cite{cslstm}, (4) DD-RCNN\cite{lu2020rcnn}, (5) ViTConv\cite{li2022distracted}, (6) Inception+ResNet+HRNN\cite{alotaibi2020distracted}, (7) LWANet\cite{lin2022lightweight}\footnote{LWANet is not compared on the AUCDD dataset, as it does not adhere to the split-by-driver setting.}, and (8) DDR-ViT-finetuned\cite{9661254}.

\begin{table}[t]
\centering
\caption{Comparison between ViT-DD with several state-of-the-art methods. The best method among each setting is highlighted in \textbf{bold}. $\downarrow$ indicates lower is better. $\uparrow$ indicates higher is better. $^*$ Results from the original papers.}
\label{tab:sota}
\resizebox{1.0\linewidth}{!}{
\begin{tabular}{l|l|cc}
    \toprule
    Experiment & Method & Accuracy $(\uparrow)$ & NLL $(\downarrow)$\\
    \midrule
    \multirow{4}{*}{AUCDD} 
    &Ensemble$^*$\cite{eraqi2019driver} & 0.9006 & 0.6400 \\
    &ADNet$^*$\cite{xiao2022attention} & 0.9022 & $-$ \\
    &C-SLSTM$^*$\cite{cslstm} & 0.9270 & 0.2793 \\
    &\textbf{ViT-DD (ours)} & \textbf{0.9359} & \textbf{0.2399} \\
    \midrule
    \multirow{2}{*}{\shortstack[l]{SFDDD \\(Split-by-Driver)}}
    &DD-RCNN$^*$\cite{lu2020rcnn} & 0.8600 & \textbf{0.3900} \\
    &\textbf{ViT-DD (ours)} &\textbf{0.9251} & 0.3972\\
    \midrule
    \multirow{5}{*}{\shortstack[l]{SFDDD \\(Split-by-Image)}} 
    & DDR-ViT-ft$^*$~\cite{9661254} & 0.9750 & $-$ \\
    & ViTConv$^*$~\cite{li2022distracted} & 0.9790 & 0.0800\\
    & Inc+Res+HRNN$^*$~\cite{alotaibi2020distracted} & 0.9930 & $-$ \\  
    & LWANet$^*$~\cite{lin2022lightweight} & 0.9937  & 0.0260  \\
    & \textbf{ViT-DD (ours)} & \textbf{0.9963} & \textbf{0.0171}\\
    \bottomrule
\end{tabular}
}
\end{table}


\subsection{Implementation Details}
The AdamW optimizer \cite{adamw} was used with weight decay set to $0.1$ for all experiments. The base learning rates were set to $0.0003$ and $0.0006$ for the SFDDD and AUCDD datasets, respectively. The learning rate was warmed up for 5 epochs, starting with a learning rate of $10^{-6}$ and then decaying to $0$ using the cosine scheduler.

To prepare the data, the driver's images were resized to $224 \times 224$ and the face images to $32 \times 32$. A patch size of $16 \times 16$ was used, and a total of $200$ patches were extracted. The ViT-DD model achieved an inference FPS of 15.56 with this input size on an NVIDIA 3090ti GPU, which is suitable for real-time applications.

For data augmentation, the Simple Random Crop strategy introduced by Touvron \etal \cite{deit3} with random horizontal flip was used. For the SFDDD dataset, the 3-Augment introduced in \cite{deit3} was also applied.

As the datasets used in this study are not very large, only the multi-head self-attention layers in the Transformer encoder were fine-tuned, as suggested by Touvron \etal \cite{touvron2022three}. The ViT-DD model was trained for 20 epochs on 1 NVIDIA A100 GPU with a batch size of 256.

\subsection{Comparison with State-of-the-Art}
The results of the performance comparisons between the proposed ViT-DD and the state-of-the-art approaches are presented in Table \ref{tab:sota}. It is worth noting that the accuracy scores are calculated in an unbiased manner that treats all categories equally, regardless of their size. Based on the results, we have the following observations:

\textit{(1)} Splitting-by-image produces a significant correlation between training and testing data. In comparison to other state-of-the-art results on SFDDD with this setting, such as LWANet\cite{lin2022lightweight}, ViT-DD achieves an accuracy improvement of $0.26\%$. This demonstrates the excellent fitting capability of ViT-DD.

\textit{(2)} When adopting a more challenging and realistic split strategy, namely, separate by driver, ViT-DD can respectively obtain $6.5\%$ and $0.9\%$ performance gains over the reported state-of-the-art results, i.e., DD-RCNN\cite{lu2020rcnn} on SFDDD and C-SLSTM\cite{cslstm} on AUCDD. This result shows the superior generalization ability of ViT-DD. The performance improvements benefit from the advantages of ViT-DD. 
First the state-of-the-art ViT is employed as the backbone network, which can provide high generalization performance if pretrained on a large-scale dataset\cite{tran2022plex}. Second, the novel multi-task multi-modal self-training method enables ViT-DD to leverage additional inductive information provided by the training signals for recognizing the emotion state of drivers, thereby improving generalization performance.

\subsection{Ablation Study}
\begin{figure}[t]%
\centering
\subfigure[Standard ViT]{%
\label{fig:ens}%
\includegraphics[width=0.9\linewidth]{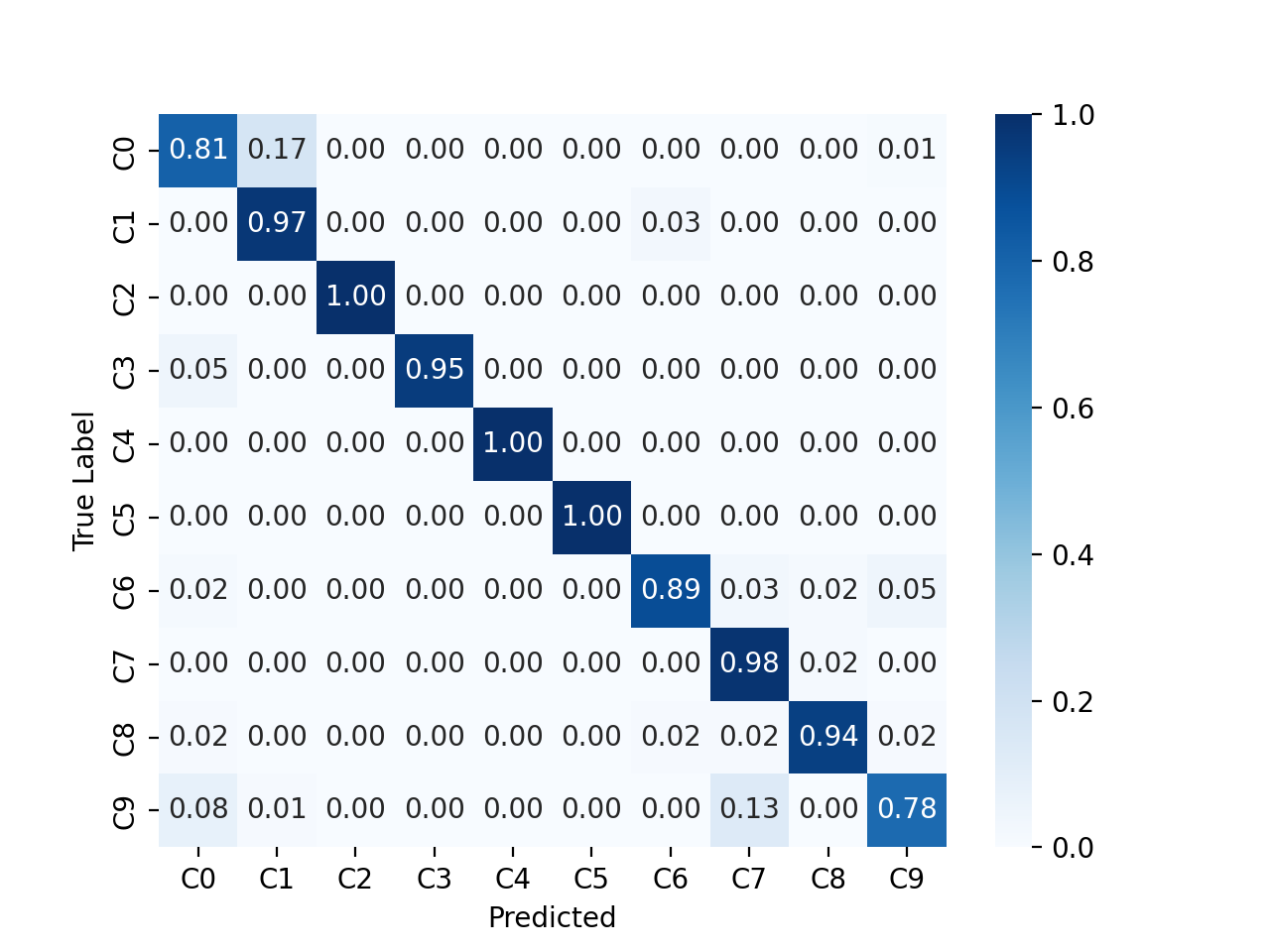}
}%
\hspace{1cm}
\vspace{-3mm}\\
\subfigure[ViT-DD]{%
\label{fig:due}%
\includegraphics[width=0.9\linewidth]{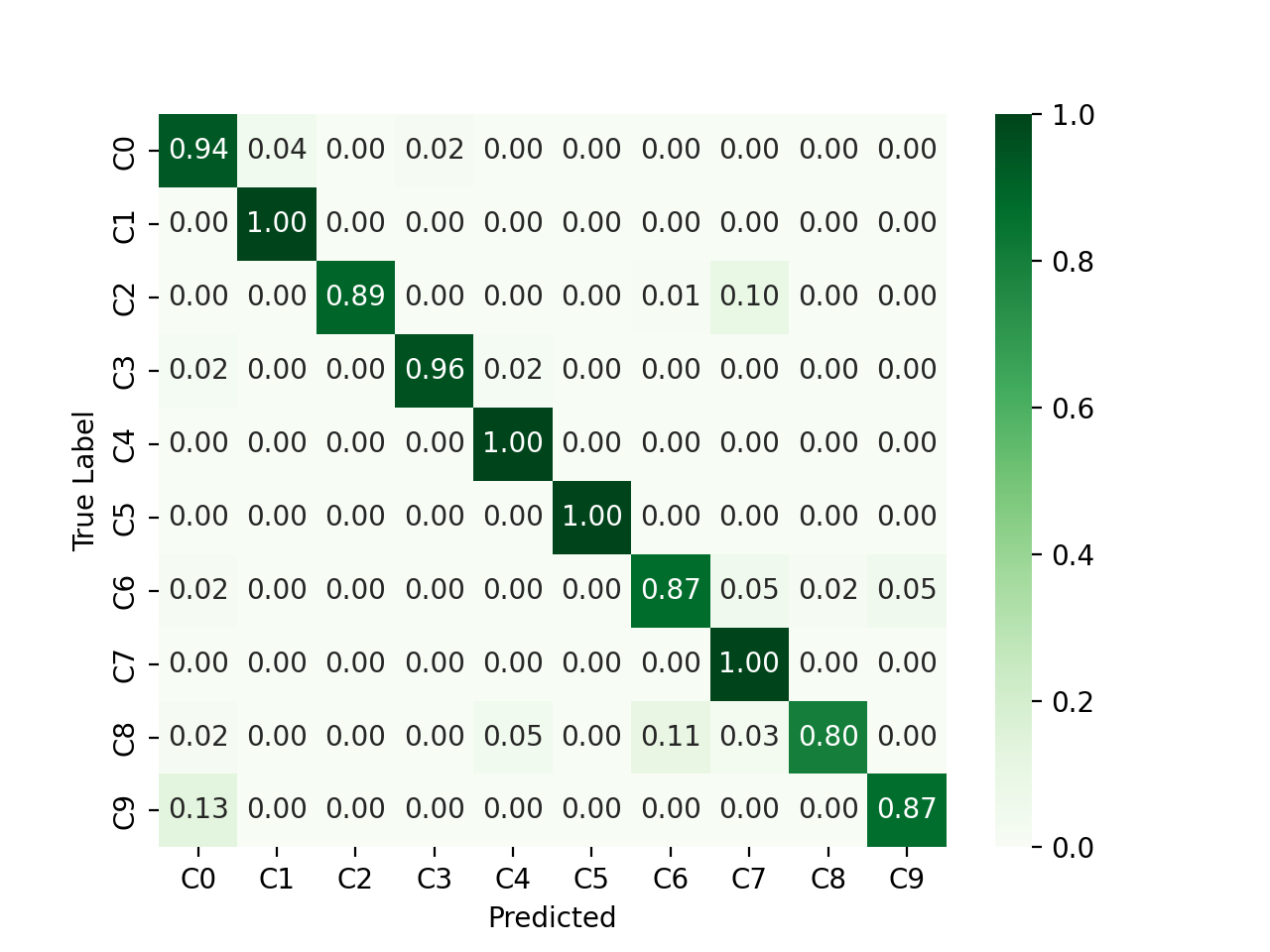}
}\\
\caption{Confusion matrices of the standard ViT and ViT-DD on AUCDD}
\label{fig:cf_matrix}
\end{figure}

The proposed ViT-DD is trained utilizing a novel multi-task multi-modal self-training procedure. To further validate the effectiveness of this strategy, an ablation study is conducted in comparison to the standard ViT trained with supervised driver distraction detection labels. The average accuracy and NLL on SFDDD split-by-driver and AUCDD datasets are shown in Table \ref{tab:ablation}. The confusion matrices on the AUCDD dataset is shown in Fig. \ref{fig:cf_matrix}. 

\begin{table}[tb]
\caption{Performance comparison between the standard ViT and the proposed ViT-DD.}
\centering
\begin{tabular}{l|c|cc}
    \toprule
    Dataset & Method & Accuracy $(\uparrow)$ & NLL $(\downarrow)$\\
    \midrule
    \multirow{2}{1.5cm}{SFDDD}  & Standard ViT\cite{visiontransformer} & 0.9036 & 0.5355 \\
    & ViT-DD & \textbf{0.9251} & \textbf{0.3972} \\
    \midrule
    \multirow{2}{1.5cm}{AUCDD} & Standard ViT\cite{visiontransformer} & 0.9092 & 0.2895 \\
    & ViT-DD & \textbf{0.9359} & \textbf{0.2399}\\
    \bottomrule
\end{tabular}
\label{tab:ablation}
\end{table}

From Table \ref{tab:ablation}, it is clear that for both datasets, ViT-DD performs better. The average accuracy improvements of ViT-DD over the standard ViT are $2.2\%$ and $2.7\%$ on the SFDDD and AUCDD datasets, respectively. This demonstrates that ViT-DD successfully leverages additional sources of information from the emotion recognition to improve the performance of learning on the task of distraction detection. From the confusion matrices, we have the following observations:

\textit{(1)} ViT-DD performs significantly better in detecting \textit{safe driving} (C0) and \textit{talking to passenger} (C9) with $13\%$ and $9\%$ increases in accuracy, respectively. This is because certain emotion states correlate strongly with these two driving behaviors. Specifically, in most cases, drivers have a neutral emotion when driving safely and tend to be happy when talking to passengers. Standard ViT tends to misclassify \textit{safe driving} as \textit{phone right} (C1). This can be resolved with the support of drivers' emotion information, as talking on the phone corresponds to all kinds of emotion status, not just neutral. Also, in standard ViT, $13\%$ of talking to passenger scenarios are misclassified as \textit{hair or makeup} (C7), compared to $0\%$ in ViT-DD due to the inclusion of emotion information.

\textit{(2)} However, ViT-DD suffers a performance loss when detecting \textit{phone left} (C2) and \textit{reaching behind} (C8). Specifically, \textit{phone left} is occasionally interpreted as \textit{hair or makeup} (C7), and \textit{reaching behind} is occasionally interpreted as \textit{drinking} (C6). In both of these cases, emotion information may mislead the detection of driver distractions: The driver's emotion state can vary when phoning, so emotion cannot provide useful information for detecting this behavior; For reaching behind, it is difficult to identify the driver's emotion, so the emotion data may not be accurate. 

\textit{(3)} It is worth noting that ViT-DD has a much higher detection accuracy on \textit{phone right} (C1) than \textit{phone left} (C2), which is due to the bias present in the dataset. The dash-board camera is positioned in front of the passenger's seat and photographs the driver from the right-hand side. As a result, detection of the phone on the right is much simpler than on the left, since the phone on the right is completely visible.

\subsection{Visualization}
In order to show the interpretability of our model, the attention maps during inference on the AUCDD dataset are visualized in Fig. \ref{fig:attn_map}, which shows the interactions between the distraction token and visual tokens of various Transformer encoder layers. The attention scores are used to generate the attention maps. For visualization purposes, the 1D sequence of attention scores is reshaped according to their original spatial positions in the driver or face images. 

As seen in Fig.~\ref{fig:attn_map}, as the network becomes deeper, the distraction token gathers more precise local cues rather than the whole driver or face image signals.  In the first few layers, the whole in-cabin scene provides interference cues, but a well-trained ViT-DD can gradually concentrate on critical areas of input images. For instance, in the first \textit{safe driving} scenario, the model successfully focuses on the driving wheel region of the driver's image, which is the most informative area of the whole picture. For the second \textit{phone left} scenario, the model effectively pays the most attention to the phone region. For both face images, the model attends to the eye region, which is the most distinguishable part of the face for recognizing the facial expressions of drivers.

\section{Conclusion and Future Work} \label{sec:Con}
In this paper, a pure Transformer architecture-based method for detecting driver distractions is proposed. The developed ViT-DD trained with the novel pseudo-labeled multi-task learning algorithm can leverage information from emotion recognition to improve the performance of learning on distraction detection. Extensive experiments conducted on SFDDD and AUCDD benchmarks with the challenging split-by-driver setting demonstrate that ViT-DD achieves $6.5\%$ and $0.9\%$ performance improvements as compared to the best state-of-the-art driver distraction detection approaches. As the next step, additional training signals available from in-cabin camera, such as gaze tracking and head pose tracking, can be incorporated into distraction detection or other driver behavior prediction tasks.

\bibliographystyle{IEEEtran}
\bibliography{ref,tiv,ma}

\end{document}